%% file: main.tex
\documentclass{article}

 \usepackage[main, final]{neurips_2026}

\usepackage[utf8]{inputenc} %
\usepackage[T1]{fontenc}    %
\usepackage{hyperref}       %
\usepackage{url}            %
\usepackage{booktabs}       %
\usepackage{amsfonts}       %
\usepackage{nicefrac}       %
\usepackage{microtype}      %
\usepackage{xcolor}         %
\hypersetup{
  colorlinks=true,
  linkcolor=red!50!black,
  citecolor=blue!50!black,
  urlcolor=magenta!50!black,
  pdftitle={CP-MLPs: A Tensor-Rank Theory of Tied and Untied MLP Blocks},
  pdfauthor={Md Rifat Arefin, Farzaneh Heidari, Irina Rish, Guillaume Rabusseau}
}

\usepackage{amsmath,amssymb,amsthm,mathtools,bm}
\usepackage{array}
\usepackage{multirow}
\usepackage{tikz}
\usetikzlibrary{arrows.meta,positioning,calc,decorations.pathreplacing}
\usepackage[most]{tcolorbox}
\usepackage{enumitem}

\DeclareMathOperator{\Sym}{Sym}

\title{CP-MLPs: A Tensor-Rank Theory of Tied and Untied MLP Blocks}

\author{ Md Rifat Arefin \quad Farzaneh Heidari \quad Irina Rish \quad Guillaume Rabusseau \\ Mila -- Quebec AI Institute \\ Universit\'e de Montr\'eal \\ Montr\'eal, QC, Canada \\ \texttt{rifat.arefin@mila.quebec} }

\newtheorem{theorem}{Theorem}[section]
\newtheorem{proposition}[theorem]{Proposition}
\newtheorem{lemma}[theorem]{Lemma}
\newtheorem{corollary}[theorem]{Corollary}
\newtheorem{definition}[theorem]{Definition}
\newtheorem{remark}[theorem]{Remark}

\newcommand{\Wrk}{\operatorname{Wr}}

\tcbset{
  enhanced,
  boxrule=0.8pt,
  arc=2mm,
  left=2mm,
  right=2mm,
  top=1.5mm,
  bottom=1.5mm,
  colback=white,
  colframe=black!65
}

\newtcolorbox{keybox}[1][]{colback=blue!4,colframe=blue!55!black,title=#1}
\newtcolorbox{storybox}[1][]{colback=green!4,colframe=green!45!black,title=#1}
\newtcolorbox{warnbox}[1][]{colback=orange!5,colframe=orange!70!black,title=#1}
\newtcolorbox{hypbox}[1][]{colback=purple!4,colframe=purple!55!black,title=#1}
\newtcolorbox{expbox}[1][]{colback=red!3,colframe=red!55!black,title=#1}

\renewcommand{\acksection}{\section*{Acknowledgments}}

\begin{document}

\maketitle

\input{sections/0.abstract}

\input{sections/1.introduction}
\input{sections/2.related_works}

\input{sections/3.methodology}

\input{sections/4.expresivity}

\input{sections/5.inductive_bias}
\input{sections/7.experimental_setup}

\input{sections/8.results}
\input{sections/9.conclusions}

\begin{ack}
We acknowledge the support from the Canada CIFAR AI Chair Program and from the Canada Excellence Research Chairs Program. The research was enabled in part by computational resources provided by the Digital Research Alliance of Canada and Mila-Quebec AI Institute.
\end{ack}

\bibliographystyle{plainnat}
\bibliography{references}

\appendix

\input{sections/appendix}

\newpage

\end{document}

%% file: sections/0.abstract.tex
\begin{abstract}
Transformer feed-forward blocks combine several design choices: squared
activations, gating, and tied versus untied parameter sharing, whose separate
contributions to expressivity remain poorly understood. We introduce
\emph{CP-MLPs}, a tensor framework in which each multiplicative
unit contributes a rank-1 interaction with three roles: detecting an input
feature, reading a value, and writing to the output. Tied gated blocks such as ReLU$^2$ force the detector and value roles to share the same direction; untied gated blocks such as ReGLU and SwiGLU let them differ. We show that this distinction yields exact, degree-wise tensor-rank separations: untied detector-value units are strictly more width-efficient for indefinite quadratic interactions and role-asymmetric monomials, while tied units are structurally matched to symmetric pure-power targets. 
{The width advantage of untying therefore depends on the target structure.
Experiments show loss plateaus below the exact-representation width, but these plateaus alone do not establish a positive approximation gap: arbitrarily accurate approximation is governed by border rank.}
In trained gated MLPs, untied blocks retain an advantage in selected comparisons at matched input-projection budget.
Code is available at \url{https://github.com/rarefin/CP-MLPs}.
\end{abstract}

%% file: sections/1.introduction.tex
\section{Introduction}
\label{sec:introduction}

Feed-forward layers account for roughly two-thirds of the parameters in a transformer~\citep{Vaswani2017AttentionIA} language model~\citep{geva2021transformer}, and a wide range of designs have been proposed for them. Common variants include squared activations such as $\mathrm{ReLU}^2$~\citep{so2021searching}, gated blocks such as ReGLU and SwiGLU~\citep{shazeer2020glu}, 
{and tied variants of SwiGLU}. Bilinear MLPs are less common as training blocks in modern LLMs, but they provide a clean quadratic base case and have received attention in the interpretability literature~\citep{pearce2024bilinear}. These architectures differ not only in their gating nonlinearity, but also in whether the same direction is used both to detect a feature and to read out its value. In this work we ask when untying these two roles improves expressivity, in the sense of reducing the number of channels required to represent a given interaction.

{$\mathrm{ReLU}^2$ and tied SwiGLU} use the same direction for the detector and value. Untied blocks such as bilinear MLPs, ReGLU, and SwiGLU allow the two to differ. This decouples two design choices that are typically conflated: the choice of gate, and whether the detector and value branches are tied.

The comparison is naturally made after input symmetrization. A polynomial function evaluates its coefficient tensor on repeated copies of the same input, so only the input-symmetric part of the tensor contributes to the function it represents~\citep{comon2008symmetric}. Tied and untied blocks therefore parameterize the same space of input-symmetric tensors, but through 
{different families of atoms}. Untying helps when the target interaction is better matched to an untied detector-value atom than to a tied pure-power atom.

Our analysis gives exact width separations at fixed polynomial degree. At degree two, untied bilinear atoms can represent certain indefinite quadratic interactions with strictly fewer channels than tied atoms. At higher degree, role-asymmetric monomials are realized by a single untied detector-value atom but require multiple tied pure-power atoms. For smooth gates such as SwiGLU, these statements transfer degree by degree to the Taylor coefficient tensors.{ These degree-wise results give necessary, but not generally sufficient, width bounds for the full analytic function.} The point is not that tying is uniformly worse: tied atoms are well suited to symmetric pure-power targets, and untied atoms to interactions that separate detection from value.

\par
{Our coefficient-space experiments distinguish exact rank, border rank, and numerical recovery (Appendices~\ref{app:extra_diagnostics}--\ref{app:extra_random}). A positive approximation gap requires width below border rank, the minimum width permitting arbitrarily accurate approximation. Below exact rank but at or above border rank, the error infimum is zero despite possible numerical loss plateaus. Both families represent pure-power targets at width one, so differences there concern optimization. The untied advantage also appears in selected trained gated-MLP comparisons under parameter matching.
\par}

\paragraph{Contributions.}
\begin{enumerate}[leftmargin=1.5em]
    \item \textbf{A unified CP-MLP framework.}
    We express bilinear,
    {$\mathrm{ReLU}^2$, tied SwiGLU, ReGLU}, and SwiGLU blocks in a shared atom-based tensor notation with detector, value, and output roles.
    \item \textbf{Exact atom-rank separations.}
    We prove that untied detector-value atoms achieve strictly lower width on indefinite quadratic interactions and role-asymmetric monomials, while tied atoms remain well matched to pure-power targets.
    \item 
    \textbf{Empirical link between exact rank, border rank, and loss.}
{We distinguish positive approximation gaps from zero-infimum limits and optimization failures, and evaluate tied and untied gated MLPs under width and parameter-budget controls.
\par}
\end{enumerate}

%% file: sections/2.related_works.tex
\section{Related Work}
\label{sec:related_work}

\paragraph{Transformer MLP variants and interpretations.}
Transformer feed-forward designs have diversified considerably. GLU-style activations were introduced for language modeling by \citet{dauphin2017language}, and variants such as ReGLU and SwiGLU were later shown to improve Transformer feed-forward layers~\citep{shazeer2020glu}. Squared-ReLU activations emerged as effective Transformer primitives through architecture-search studies~\citep{so2021searching}, building on a longer line of activation-function design that includes Swish/SiLU~\citep{ramachandran2017searching}. In parallel, interpretability work has argued that Transformer feed-forward layers behave like structured key-value memories rather than generic nonlinear post-processing~\citep{geva2021transformer}. These works motivate studying MLP blocks as structured computational modules,
{but do not directly isolate the expressivity effect} of tying versus untying detector and value branches, which is the focus of this paper.

\paragraph{Tensor, polynomial, and Waring-rank views.}
Our framework uses canonical polyadic (CP) tensor decomposition, where tensors are written as sums of rank-1 outer products~\citep{kolda2009tensor}. Recent work on bilinear MLPs also exploits tensor structure to analyze feed-forward blocks, observing that bilinear layers admit a third-order tensor representation~\citep{pearce2024bilinear}. The $\Pi$-Net family of \citet{chrysos2020p,chrysos2021deep}  parameterizes polynomial expansions of the input directly through tensor decompositions, demonstrating that such polynomial blocks are competitive with standard activations on a range of tasks. Polynomial networks have also been studied with algebraic and tensor tools, for example through the geometry of polynomial function classes~\citep{kileel2019expressive}. Our work differs in that we do not propose polynomial networks as an architecture; we use polynomial structure as an analytical lens for comparing existing MLP variants.

For the higher-degree separation we rely on classical Waring rank: a scalar tied degree-$k$ channel is a pure power $(\ell^\top x)^k$, so tied atom rank coincides with Waring rank. The monomial Waring-rank formula of \citet{carlini2011solution}, together with the real-rank result of \citet{carlini2017real}, implies that $x_1^{k-1} x_2$ requires $k$ tied atoms but only one untied detector-value atom.{ This exact-rank separation differs from the approximation threshold: the same monomial has tied border rank two~\citep{comas2011rank}.} Our contribution is to connect these rank facts to practical MLP atom families, showing that tied and untied blocks draw on different atoms inside the same input-symmetric tensor space.

%% file: sections/3.methodology.tex
\section{The CP-MLP framework}
\label{sec:cp_mlp}

We use a common tensor notation for the multiplicative feed-forward blocks studied in this paper. Bilinear blocks multiply two linear projections, GLU-style blocks gate one projection by another, and square-style blocks reuse the same projection twice. In every case, each hidden channel produces a product of projected input directions, and the output matrix sums these channel-wise products. 
{At each active degree, the raw coefficient tensor has the structure of a canonical polyadic (CP) decomposition}~\citep{kolda2009tensor}, and we call the resulting representation a \emph{CP-MLP}.

The framework separates two design choices: \emph{the gate}, which selects the polynomial degrees that appear in a local expansion, and \emph{the tying pattern}, which selects whether the detector and value directions of a channel are shared. The expressivity analysis in Section~\ref{sec:expressivity_cp_rank} isolates the effect of the second choice.

\begin{definition}[CP-MLP block]
\label{def:cp_mlp}
Fix a scalar gate \(g:\mathbb R\to\mathbb R\) and let
\(G:\mathbb R^{m}\to\mathbb R^{m}\) be its coordinatewise extension,
\(G(z)=\bigl(g(z_1),\dots,g(z_m)\bigr)^\top\).
Given weights \(U,V\in\mathbb R^{m\times d}\) (rows \(u_a^\top,v_a^\top\)) and
\(W\in\mathbb R^{p\times m}\) (columns \(w_a\)), a width-\(m\)
CP-MLP block is
\[
    F(x)
    =
    W\bigl(G(Ux)\odot Vx\bigr)
    =
    \sum_{a=1}^{m}
    w_a\,g(u_a^\top x)\,(v_a^\top x).
\]
We call \(u_a\) the \emph{detector direction}, \(v_a\) the \emph{value
direction}, and \(w_a\) the \emph{output direction} of channel \(a\).
{A \emph{tied} block constrains \(V=U\); an \emph{untied} block parameterizes \(U\) and \(V\) separately, allowing equality as a special case. We consider bias-free blocks throughout.}
\end{definition}

The detector direction selects the scalar feature passed through the gate, while the value direction supplies the linear signal that the gate modulates. Tying forces both roles onto the same direction; untying lets them differ. Figure~\ref{fig:cp_mlp_atom} shows the corresponding three-leg view of a single channel.

\input{figures/nccp_figure}

\subsection{Degree-wise atoms}
\label{sec:degreewise_cp_atoms}
Assume the gate is analytic at the origin, \(g(z)=\sum_{j\ge 0}c_j z^j\)
(piecewise-analytic gates such as \(\operatorname{ReLU}\) are treated
region-wise; see Table~\ref{tab:cp_mlp_variants}). Substituting \(z=u_a^\top x\)
and multiplying by \(v_a^\top x\), one channel contributes
\(\sum_{j\ge0}c_j(u_a^\top x)^{j}(v_a^\top x)\); reindexing by total degree
\(k=j+1\), the degree-\(k\) term carries the coefficient \(c_{k-1}\). Writing
\(T[x,\ldots,x]\in\mathbb R^p\) for the contraction of the \(k\) input modes of
\(T\in\mathbb R^p\otimes(\mathbb R^d)^{\otimes k}\) against \(x\) (output mode
free; conventions in Appendix~\ref{app:tensors_cp}), the degree-\(k\) component
of the block is \(F^{(k)}(x)=T^{(k)}[x,\ldots,x]\), with raw coefficient tensor
\[
    T^{(k)}
    =
    c_{k-1}\sum_{a=1}^{m}
    w_a\otimes u_a^{\otimes(k-1)}\otimes v_a
    \in
    \mathbb R^p\otimes(\mathbb R^d)^{\otimes k}.
\]
Each active degree thus admits a CP decomposition with at most \(m\) rank-one
terms, one per channel. We define \(\mathcal{S}(g)\) for the \emph{polynomial
signature} of the gate, i.e.\ the set of active degrees.
\begin{definition}[Polynomial signature]
\label{def:polynomial_signature}
The polynomial signature of \(g\) is \(\mathcal{S}(g):=\{k\ge 1: c_{k-1}\neq 0\}\).
\end{definition}
The gate selects which degrees are active through \(\mathcal{S}(g)\); the tying pattern selects the atom used at each active degree. When \(v_a=u_a\), the raw degree-\(k\) atom collapses to \(w_a\otimes u_a^{\otimes k}\); when \(v_a\) is independent of \(u_a\), it retains the detector-value form \(w_a\otimes u_a^{\otimes(k-1)}\otimes v_a\).

\subsection{Effective atoms after symmetrization}
\label{sec:effective_atoms}

The tensor above distinguishes detector and value slots, but the function it represents evaluates the tensor on repeated copies of the same input, so input-slot order is not observable. We write \(\Sym(T)\) for the symmetrization of \(T\), the average of \(T\) over all permutations of its input modes (with the output mode, when present, left fixed), then
\[
    T[x,\ldots,x]=\Sym(T)[x,\ldots,x].
\]
The proof is immediate: every permutation of \(T\) receives the same vector \(x\) in every input slot. A full statement, together with the quadratic matrix specialization, is given in Appendix~\ref{app:cp_mlp_symmetrization}.

\par
{After symmetrization, each architecture yields a family of input-symmetric atoms; an untied atom can have tensor rank greater than one.
\par}

\begin{definition}[Effective tied and untied CP-MLP atoms]
\label{def:effective_atoms}
The degree-\(k\) effective untied atom is
\[
    \mathcal A_{\mathrm{untied}}^{(k)}(w,u,v)
    :=
    w\otimes
    \Sym\bigl(u^{\otimes(k-1)}\otimes v\bigr)
    \in
    \mathbb R^p\otimes\Sym^k(\mathbb R^d),
\]
where $\Sym^k(\mathbb R^d)$ denotes the space of symmetric order-$k$ tensors (Definition~\ref{def:input_symmetric}). The degree-\(k\) effective tied atom is the specialization \(v=u\):
\[
    \mathcal A_{\mathrm{tied}}^{(k)}(w,u)
    :=
    w\otimes u^{\otimes k}
    \in
    \mathbb R^p\otimes\Sym^k(\mathbb R^d).
\]
\end{definition}

Tied and untied blocks therefore parameterize tensors in the same symmetric space \(\mathbb R^p\otimes\Sym^k(\mathbb R^d)\), but draw on different atom families: tied blocks use pure-power atoms \(w\otimes u^{\otimes k}\), while untied blocks use symmetrized detector-value atoms \(w\otimes\Sym(u^{\otimes(k-1)}\otimes v)\). Setting \(v=u\) recovers the tied family as a special case. The benefit of untying is not that antisymmetric components become visible, but that the symmetrized untied family can represent certain symmetric tensors with strictly fewer atoms.

\begin{table}[t]
\centering
\caption{
Feed-forward blocks as CP-MLPs. The gate controls which degrees are active through its polynomial signature; the tying pattern controls whether detector and value directions are shared.
}
\label{tab:cp_mlp_variants}
\small
\setlength{\tabcolsep}{4pt}
\begin{tabular}{@{}lllll@{}}
\toprule
\textbf{Block}
& \textbf{Gate \(g(z)\)}
& \textbf{Tying}
& \textbf{Signature \(\mathcal{S}(g)\)}
& \textbf{Effective atom} \\
\midrule
Bilinear~\citep{pearce2024bilinear}
& \(z\)
& untied
& \(\{2\}\)
& \(w\otimes\Sym(u\otimes v)\) \\
ReGLU~\citep{shazeer2020glu}
& \(\operatorname{ReLU}(z)\)
& untied
& \(\{2\}\)\(^{\dagger}\)
& region-wise \(w\otimes\Sym(u\otimes v)\) \\
SwiGLU~\citep{shazeer2020glu}
& \(\operatorname{SiLU}(z)=z\sigma(z)\)
& untied
& \(\{2,3,5,7,\ldots\}\)
& \(w\otimes\Sym(u^{\otimes(k-1)}\otimes v)\) \\
\addlinespace
ReLU\(^2\)~\citep{so2021searching}
& \(\operatorname{ReLU}(z)\)
& tied
& \(\{2\}\)\(^{\dagger}\)
& region-wise \(w\otimes u^{\otimes 2}\) \\
{Tied SwiGLU} & {\(\operatorname{SiLU}(z)=z\sigma(z)\)} & {tied} & {\(\{2,3,5,7,\ldots\}\)} & {\(w\otimes u^{\otimes k}\)} \\
\bottomrule
\end{tabular}
\par\smallskip
{\footnotesize \(^{\dagger}\)Region-wise: ReLU is piecewise linear, so the channel is a degree-2 polynomial inside each fixed activation region.}
\end{table}

Table~\ref{tab:cp_mlp_variants} places the blocks studied in this paper along the two axes of the framework: the gate and the tying pattern.
Bilinear, ReGLU, and SwiGLU differ only in their gates and share the untied detector-value atom structure. 
{ReLU\(^2\) and tied SwiGLU use tied atoms, with the same direction serving both the detector and value roles. Tied SwiGLU sets \(v=u\) while retaining the SiLU gate, giving \(\operatorname{SiLU}(z)z=z^2\sigma(z)\).
\par} The next section asks how many atoms of each family are required to represent a given target tensor.

%% file: figures/nccp_figure.tex
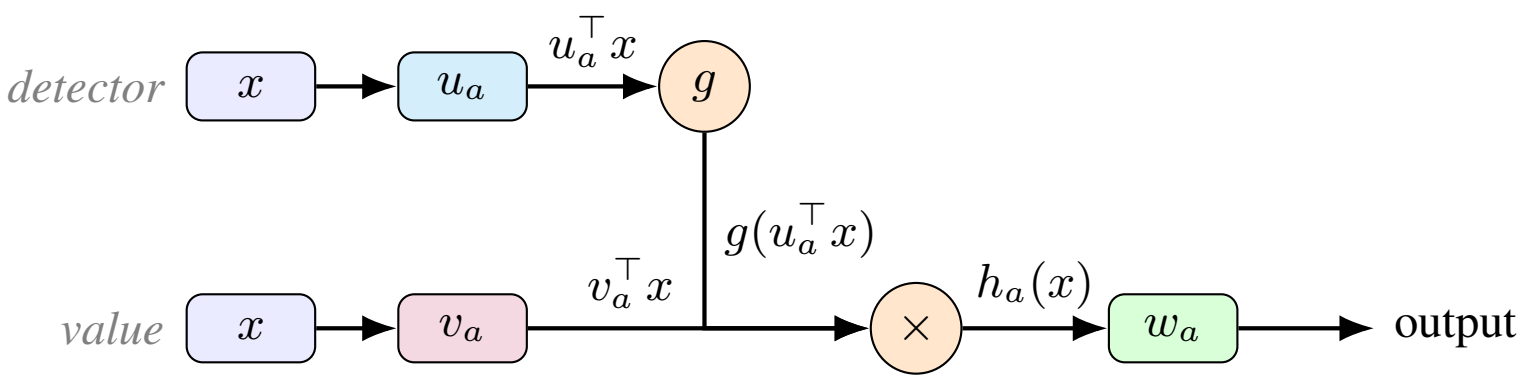
\begin{figure}[t]
\centering
\begin{tikzpicture}[
    >=Latex,
    every node/.style={font=\small},
    box/.style={draw, rounded corners=3pt, minimum width=0.85cm,
                minimum height=0.45cm, inner sep=2pt},
    op/.style={draw, circle, minimum size=0.6cm, inner sep=0pt},
    arr/.style={->, thick, >=Latex},
    rolelabel/.style={font=\small\itshape, text=gray}
]

    \node[box, fill=blue!8]   (x1) at (-3.4,  0.8) {$x$};
    \node[box, fill=blue!8]   (x2) at (-3.4, -0.8) {$x$};

    \node[box, fill=cyan!15]  (u)  at (-2.0,  0.8) {$u_a$};
    \node[box, fill=purple!15](v)  at (-2.0, -0.8) {$v_a$};

    \node[op,  fill=orange!20] (g)   at (-0.4,  0.8) {$g$};
    \node[op,  fill=orange!20] (mul) at ( 1, -0.8) {$\times$};

    \node[box, fill=green!15] (w)   at ( 2.7, -0.8) {$w_a$};

    \node[rolelabel, anchor=east] at (-3.85,  0.8) {detector};
    \node[rolelabel, anchor=east] at (-3.85, -0.8) {value};

    \draw[arr] (x1) -- (u);
    \draw[arr] (u) -- node[above]{$u_a^\top x$} (g);
    \draw[arr] (g.south) -- node[right]{$g(u_a^\top x)$}
                         (g.south |- mul.west) -- (mul.west);

    \draw[arr] (x2) -- (v);
    \draw[arr] (v.east) -- node[above, pos=0.3]{$v_a^\top x$} (mul);

    \draw[arr] (mul) -- node[above]{$h_a(x)$} (w);
    \draw[arr] (w.east) -- ++(0.9, 0) node[right]{output};
\end{tikzpicture}
\caption{%
\textbf{Leg diagram of a single CP-MLP channel.}
The channel computes $h_a(x) = g(u_a^\top x)\,(v_a^\top x)$ and writes
the result through $w_a$.  At each active polynomial degree, the channel
contributes one rank-1 CP atom with output, detector, and value legs.}
\label{fig:cp_mlp_atom}
\end{figure}

%% file: sections/4.expresivity.tex
\section{Expressivity via CP-MLP atom rank}
\label{sec:expressivity_cp_rank}

The atom families of Section~\ref{sec:effective_atoms} suggest a natural notion of width: the smallest number of atoms needed to express a given coefficient tensor. Fixing an active degree \(k\), each CP-MLP channel contributes one effective atom in \(\mathcal T_k:=\mathbb R^p\otimes \Sym^k(\mathbb R^d)\), so the question we want to answer is how many such atoms are required to represent a target \(T\in\mathcal T_k\). We call this minimum the \emph{CP-MLP atom rank}; since each atom corresponds to one channel at degree \(k\), it is the degree-wise width required to represent a coefficient tensor.

For polynomial gates, \(T^{(k)}\) is the ordinary degree-\(k\) coefficient tensor; for analytic gates, it is the degree-\(k\) Taylor coefficient tensor, satisfying \(D^k F(0)=k!\,\Sym(T^{(k)})\). Throughout this section \(k\) denotes an active degree of the gate, and any nonzero scalar Taylor coefficient is absorbed into the output factor \(w\), so the rank comparisons below depend only on the channel atom family.

\subsection{Atom-rank definitions}
\label{sec:atom_rank_definitions}

\begin{definition}[Tied CP-MLP atom rank]
\label{def:tied_cp_mlp_rank}
For \(T\in\mathcal T_k\), the \emph{tied CP-MLP atom rank} \(r_{\mathrm{tied}}(T)\) is the smallest integer \(m\) such that
\(
    T
    =
    \sum_{a=1}^{m}
    w_a\otimes u_a^{\otimes k},
    \qquad
    w_a\in\mathbb R^p,\quad u_a\in\mathbb R^d.
\)
\end{definition}

\begin{definition}[Untied CP-MLP atom rank]
\label{def:untied_cp_mlp_rank}
For \(T\in\mathcal T_k\), the \emph{untied CP-MLP atom rank} \(r_{\mathrm{untied}}(T)\) is the smallest integer \(m\) such that
\(
    T
    =
    \sum_{a=1}^{m}
    w_a\otimes
    \Sym\bigl(
        u_a^{\otimes(k-1)}\otimes v_a
    \bigr),
    \qquad
    w_a\in\mathbb R^p,\quad u_a,v_a\in\mathbb R^d.
\)
\end{definition}

Both ranks are minimum-width quantities at a fixed degree, counting the fewest atoms required to express a coefficient tensor in the corresponding family. Since a tied atom is recovered from an untied atom by setting \(v_a=u_a\), the untied family contains the tied family, which immediately implies the following inequality.

\begin{proposition}[Untied atoms contain tied atoms]
\label{prop:untied_le_tied}
For every \(T\in\mathcal T_k\), \(r_{\mathrm{untied}}(T)\le r_{\mathrm{tied}}(T).\)
\end{proposition}

\begin{proof}
Any tied representation \(T=\sum_a w_a\otimes u_a^{\otimes k}\) is also an untied representation with \(v_a=u_a\), since \(u_a^{\otimes k}=\Sym(u_a^{\otimes(k-1)}\otimes u_a)\). Minimizing over tied representations yields the claim.
\end{proof}

The remainder of the section asks when this inequality is strict. We first study degree two, where the coefficient tensor is a symmetric matrix, and then prove a higher-degree separation using a role-asymmetric monomial.

\begin{remark}[Scalar targets suffice]
\label{rem:scalar}
\label{sec:scalar_targets_suffice}
For a scalar target \(S\in\Sym^k(\mathbb R^d)\), embedding \(T=e_o\otimes S\) gives \(r_{\mathrm{tied}}(T)=r_{\mathrm{tied}}(S)\) and \(r_{\mathrm{untied}}(T)=r_{\mathrm{untied}}(S)\): output projection cannot increase atom rank, and any scalar representation lifts by tensoring with \(e_o\). Full details are in Appendix~\ref{app:scalar_projection}. The separations below are stated for scalar targets.
\end{remark}

\subsection{Degree-two bilinear atoms}
\label{sec:bilinear_case}

We first specialize to degree \(2\), where a scalar coefficient tensor is a symmetric matrix and the atom-rank question reduces to representing quadratic forms
\[
    f_M(x)=x^\top Mx,
    \qquad
    M=M^\top\in\mathbb R^{d\times d}.
\]
A tied atom contributes the signed square \(\alpha(u^\top x)^2\), corresponding to the rank-1 symmetric matrix \(\alpha u u^\top\). An untied atom contributes a product of two linear forms, \(\alpha(u^\top x)(v^\top x)\), which after symmetrization becomes
\[
    \tfrac{\alpha}{2}(uv^\top+vu^\top).
\]
A tied atom therefore contributes one quadratic direction, while an untied atom can couple two. Our first result records this coarse rank comparison.

\begin{theorem}[Coarse bilinear rank bounds]
\label{thm:bilinear_rank_bounds}
Let \(M\in\mathbb R^{d\times d}\) be symmetric. If \(M\) is represented by a tied bilinear block of width \(m\), then \(\operatorname{rank}(M)\le\min(m,d)\). If \(M\) is represented by an untied bilinear block of width \(m\), then \(\operatorname{rank}(M)\le\min(2m,d)\). Both bounds are tight.
\end{theorem}

\begin{proof}[Proof sketch]
Each tied unit contributes the rank-1 symmetric matrix \(u_a u_a^\top\), so a width-\(m\) tied block produces matrices of rank at most \(\min(m,d)\). For untied blocks, only the symmetric part of \(u_a v_a^\top\) contributes; the resulting matrix \(\tfrac12(u_a v_a^\top+v_a u_a^\top)\) has image in \(\operatorname{span}\{u_a,v_a\}\) and therefore rank at most two, and summing \(m\) such terms gives the bound. Both bounds are tight by choosing independent coordinate directions.
\end{proof}

Theorem~\ref{thm:bilinear_rank_bounds} shows that an untied atom can carry twice the rank of a tied atom, but rank alone does not determine which matrices a given width can represent. The rank-2 gain has a specific sign structure. From the difference-of-squares identity
\[
    (u^\top x)(v^\top x)
    =
    \tfrac14\bigl((u+v)^\top x\bigr)^2
    -
    \tfrac14\bigl((u-v)^\top x\bigr)^2,
\]
a non-collinear untied atom is a difference of two squares. It can pair one positive with one negative quadratic direction---representing interactions like \(a^2-b^2=(a+b)(a-b)\)---but cannot compress two same-sign directions such as \(a^2+b^2\) into a single atom. The exact width is therefore controlled not by rank, but by how many positive and negative quadratic directions the target carries.

Let \(n_+(M)\) and \(n_-(M)\) denote the numbers of positive and negative eigenvalues of \(M\).

\begin{theorem}[Exact bilinear atom ranks]
\label{thm:exact_bilinear_expressivity}
For every symmetric matrix \(M\in\mathbb R^{d\times d}\),
\[
    r_{\mathrm{tied}}(M)
    =
    \operatorname{rank}(M)
    =
    n_+(M)+n_-(M),
\]
whereas
\[
    r_{\mathrm{untied}}(M)
    =
    \max\{n_+(M),n_-(M)\}.
\]
\end{theorem}

\begin{proof}[Proof sketch]
For tied atoms, each channel contributes one signed rank-1 square, so it needs one atom per nonzero eigendirection, and the spectral decomposition achieves this number. For untied atoms, the difference-of-squares identity expresses any width-\(m\) representation as \(M=AA^\top-BB^\top\) with \(A,B\in\mathbb R^{d\times m}\). On \(\ker(A^\top)\), \(x^\top Mx=-\|B^\top x\|^2\le 0\), so \(M\) cannot be positive definite on a subspace of dimension greater than \(m\); the analogous argument on \(\ker(B^\top)\) gives the same bound for negative directions, so \(m\ge\max\{n_+(M),n_-(M)\}\). The matching upper bound is achieved by pairing opposite-sign eigendirections one atom at a time, with unpaired directions handled by the tied special cases \(u=\pm v\). Full proofs of Theorems~\ref{thm:bilinear_rank_bounds} and~\ref{thm:exact_bilinear_expressivity} are in Appendix~\ref{app:bilinear_width}.
\end{proof}

\begin{corollary}[When untied bilinear atoms help]
\label{cor:bilinear-regimes}
If \(M\succeq 0\) or \(M\preceq 0\),
\[
    r_{\mathrm{tied}}(M)=r_{\mathrm{untied}}(M)=\operatorname{rank}(M).
\]
If \(M\) is indefinite,
\[
    r_{\mathrm{tied}}(M)-r_{\mathrm{untied}}(M)=\min\{n_+(M),n_-(M)\}.
\]
Untied atoms are therefore most width-efficient on indefinite quadratic interactions; they offer no advantage on purely same-sign quadratic structure.
\end{corollary}

\paragraph{Connection to hard-gated blocks.}
The same degree-two atom gap appears regionwise in hard-gated CP-MLPs. On a fixed activation pattern \(\mu\in\{0,1\}^m\), ReGLU realizes
\(
    Q_{\mu,o}
    =
    \sum_{a=1}^{m}
    \mu_a W_{oa}
    \frac{u_a v_a^\top+v_a u_a^\top}{2}
\)
for each output coordinate \(o\), whereas \(\mathrm{ReLU}^2\) realizes \(Q_{\mu,o}=\sum_a \mu_a W_{oa}\,u_a u_a^\top\). Inside each fixed region, ReGLU has access to untied rank-2 atoms with opposite-sign structure, while \(\mathrm{ReLU}^2\) is restricted to tied rank-1 atoms. Region switching adds further piecewise expressivity; Theorem~\ref{thm:exact_bilinear_expressivity} isolates the within-region atom difference.

\subsection{Higher-degree monomial separation}
\label{sec:higher_degree_monomial}

The degree-two analysis showed that untied atoms can be more width-efficient when the target contains opposite-sign quadratic structure. At higher degree, a different kind of separation emerges that is driven not by sign, but by \emph{role asymmetry}: one direction is repeated many times in the detector slot, while another direction appears only once in the value slot.

For \(k\ge 2\), consider the binary monomial
\(
    F_k(x_1,x_2)=x_1^{k-1}x_2.
\)
This target has the exact shape of a single untied atom: taking \(u=e_1\) and \(v=e_2\) gives \((u^\top x)^{k-1}(v^\top x)=x_1^{k-1}x_2\), so one untied atom suffices. A tied degree-\(k\) atom, by contrast, has the pure-power form \((\ell^\top x)^k\), so the tied atom rank of \(F_k\) is exactly its Waring rank.

\begin{definition}[Waring rank]
\label{def:waring_rank}
Let \(\mathbb F\in\{\mathbb R,\mathbb C\}\), and let \(f:\mathbb F^d\to\mathbb F\) be a homogeneous polynomial of degree \(k\). The \emph{Waring rank} of \(f\) over \(\mathbb F\), denoted \(\Wrk_{\mathbb F}(f)\), is the smallest integer \(m\) such that
\[
    f(x)=\sum_{a=1}^{m}c_a\,\ell_a(x)^k,
\]
where \(c_a\in\mathbb F\) and each \(\ell_a\) is a linear form over \(\mathbb F\).
\end{definition}

For scalar tied CP-MLPs, the real Waring rank coincides with the tied atom rank: any tied representation \(f(x)=\sum_a c_a(u_a^\top x)^k\) with \(u_a\in\mathbb R^d\) is exactly a real Waring decomposition, so
\(
    r_{\mathrm{tied}}(f)=\Wrk_{\mathbb R}(f)
\)
for every scalar homogeneous degree-\(k\) polynomial. This is the standard correspondence between symmetric tensor rank and polynomial Waring rank~\citep{comon2008symmetric}. The following classical formula computes the right-hand side for monomials.

\begin{proposition}[Waring rank of monomials; \citealp{carlini2011solution,carlini2012coprime,carlini2017real}]
\label{prop:waring_rank_monomials}
Let \(\mu(x)=x_1^{\alpha_1}\cdots x_n^{\alpha_n}\) be a monomial of degree \(k=\sum_i\alpha_i\), with sorted exponents \(1\le\alpha_1\le\cdots\le\alpha_n\). Then
\[
    \Wrk_{\mathbb C}(\mu)
    =
    \frac{1}{\alpha_1+1}\prod_{i=1}^{n}(\alpha_i+1)
    =
    \prod_{i=2}^{n}(\alpha_i+1),
\]
and \(\Wrk_{\mathbb R}(\mu)=\Wrk_{\mathbb C}(\mu)\) whenever \(\alpha_1=1\).
\end{proposition}

The complex formula is \citep[Prop.~3.1]{carlini2012coprime}; the real-complex equality for \(\alpha_1=1\) follows from \citet{carlini2017real}. For \(F_k=x_1^{k-1}x_2\) the sorted exponent vector is \((1,k-1)\), so
\[
    \Wrk_{\mathbb R}(F_k)=\Wrk_{\mathbb C}(F_k)=(k-1)+1=k.
\]

\begin{theorem}[Degree-\(k\) monomial separation]
\label{thm:monomial_separation}
For every \(k\ge 2\),
\[
    r_{\mathrm{untied}}(F_k)=1,
    \quad
    r_{\mathrm{tied}}(F_k)=k.
\]
A single untied atom therefore represents a degree-\(k\) coefficient tensor that requires \(k\) tied atoms.
\end{theorem}

\begin{proof}
We bound the two ranks separately.

\emph{Untied rank.} Setting \(u=e_1\) and \(v=e_2\),
\(
    (u^\top x)^{k-1}(v^\top x)=x_1^{k-1}x_2=F_k(x),
\)
so a single untied atom realizes \(F_k\) and \(r_{\mathrm{untied}}(F_k)\le 1\). Since \(F_k\neq 0\), it cannot be represented with zero atoms, giving \(r_{\mathrm{untied}}(F_k)=1\).

\emph{Tied rank.} Any scalar tied degree-\(k\) representation \(F_k(x)=\sum_a c_a\,\ell_a(x)^k\) is a real Waring decomposition, so \(r_{\mathrm{tied}}(F_k)=\Wrk_{\mathbb R}(F_k)=k\) by Proposition~\ref{prop:waring_rank_monomials}.
\end{proof}

{\paragraph{Exact representation versus approximation.}
The real border rank \(\underline r_{\mathsf f}(T)\) is the
smallest width permitting arbitrarily accurate approximation
of \(T\) by real atoms from family \(\mathsf f\).
In coefficient norm, a positive approximation gap occurs
exactly when \(m<\underline r_{\mathsf f}(T)\).
For scalar quadratics, both width classes are closed, so exact
and border ranks coincide (Appendix~\ref{app:closed_quadratic}).
For \(F_k=x_1^{k-1}x_2\), however,
\(\underline r_{\mathrm{tied}}(F_k)=2\), whereas
\(r_{\mathrm{tied}}(F_k)=k\).
Thus, at \(2\le m<k\), the tied error infimum is zero but unattained.
Under any finite bound on \(\sum_a\|w_a\|\|u_a\|^k\), a positive
tied approximation gap remains below the exact rank.
Appendix~\ref{app:border_rank} gives the approximation arguments;
Appendix~\ref{app:block_family} extends the separation to sums
of independent interaction blocks.
\par}

{The degree-wise expansion gives necessary width bounds for
exact representation of the full analytic function.
If a bias-free width-\(m\) block with gate \(g\) represents an
analytic target \(F\) near the origin, then
\[
m\ge \sup_{k\in\mathcal S(g)}
r_{\mathsf f}\bigl(F^{(k)}\bigr),
\qquad
\mathsf f\in\{\mathrm{tied},\mathrm{untied}\},
\]
where \(F^{(k)}\) is the degree-\(k\) homogeneous component.
Exact representation also requires \(F(0)=0\) and vanishing
components at inactive degrees.
These conditions are not generally sufficient: all degrees
share the same channels, and the gate fixes their coefficient
ratios.
Appendix~\ref{app:analytic_width} gives the proof and an explicit
example where the full required width exceeds every individual
degree-wise rank.
\par}

%% file: sections/7.experimental_setup.tex
\section{Experiments}
\label{sec:exp}

We test whether the atom-rank separations from Section~\ref{sec:expressivity_cp_rank} appear as observable width and loss effects. Indefinite quadratics realize the degree-two gap (Section~\ref{sec:exp_quad}); role-asymmetric monomials realize the higher-degree gap (Section~\ref{sec:exp_waring}); pure-power targets show that tying can be the better match (Section~\ref{sec:exp_pure}); and gated MLPs show how the separation carries over to actual gates, including under a matched input-projection budget (Section~\ref{sec:exp_lift}). Optimization, hyperparameters, and budget checks are in Appendix~\ref{app:exp}.

\subsection{Indefinite quadratic targets}
\label{sec:exp_quad}

Theorem~\ref{thm:exact_bilinear_expressivity} predicts that untied bilinear atoms are more width-efficient than tied atoms only when the target carries both positive and negative directions. To test this, we construct
\[
    Q=\operatorname{diag}(\mathbf 1^{n_+},-\mathbf 1^{n_-},\mathbf 0^{d-n_+-n_-}),
    \qquad
    y(x)=x^\top Qx,
\]
with \(d=8\), and sweep six choices of \((n_+,n_-)\) covering definite, balanced indefinite, and imbalanced indefinite targets.

\begin{figure}[t]
\centering
\includegraphics[width=0.8\textwidth]{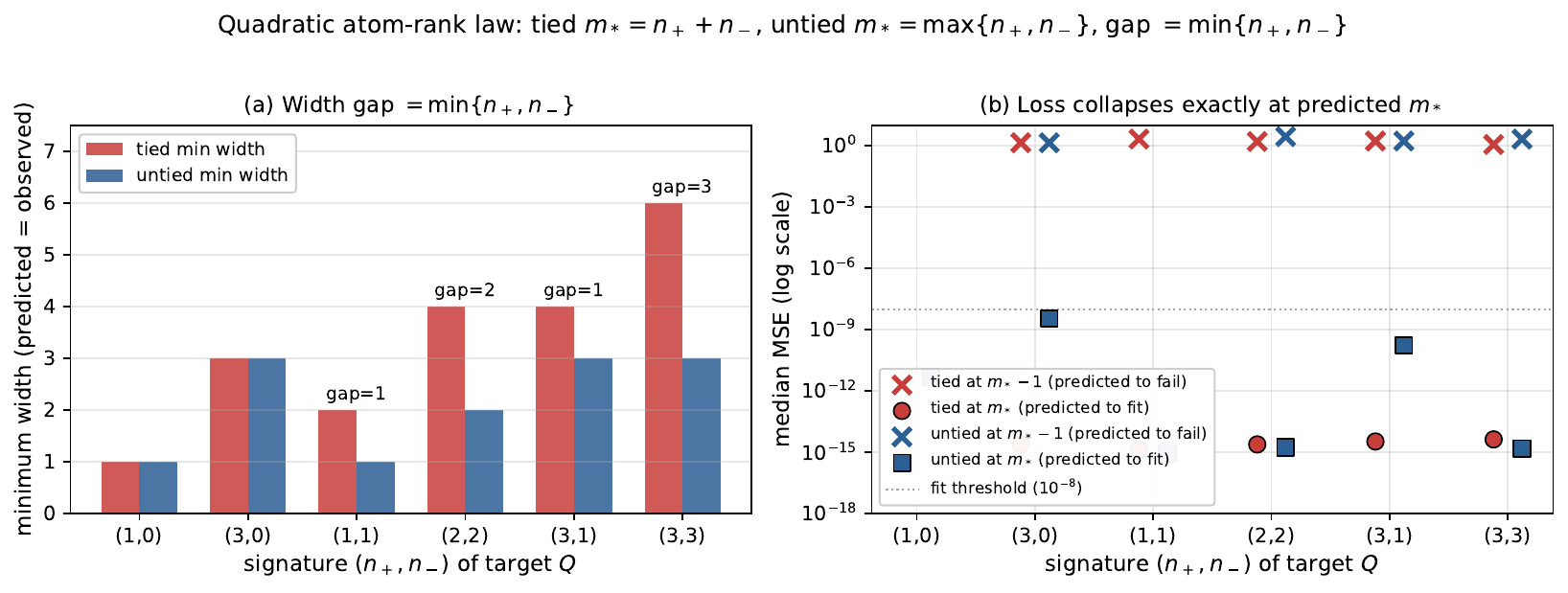}
\caption{Quadratic targets with \((n_+,n_-)\) positive and negative directions. \emph{Left:} required widths match the predictions \(n_++n_-\) (tied) and \(\max\{n_+,n_-\}\) (untied). \emph{Right:} loss stays near the target variance below the predicted width and reaches machine precision at the predicted width.}
\label{fig:quadratic}
\end{figure}

Figure~\ref{fig:quadratic} confirms the exact threshold: tied blocks require width \(n_++n_-\), untied blocks require width \(\max\{n_+,n_-\}\). At one width below the prediction, the loss stays near the target variance; at the predicted width, the loss drops to machine precision. Definite targets show no tied/untied gap, while indefinite targets show the predicted gap \(\min\{n_+,n_-\}\).

\subsection{Role-asymmetric monomial targets}
\label{sec:exp_waring}

We next test the higher-degree separation of Theorem~\ref{thm:monomial_separation}. For
\[
    F_k(x_1,x_2)=x_1^{k-1}x_2,
    \qquad
    k\in\{2,3,4,5\},
\]
the theory predicts untied atom rank \(1\) and tied atom rank \(k\). We fit the target directly in monomial coefficient space and sweep width \(m=1,\ldots,k+2\); coefficient-space fitting removes sampling noise and isolates the rank effect.

\begin{figure}[t]
\centering
\includegraphics[width=0.80\textwidth]{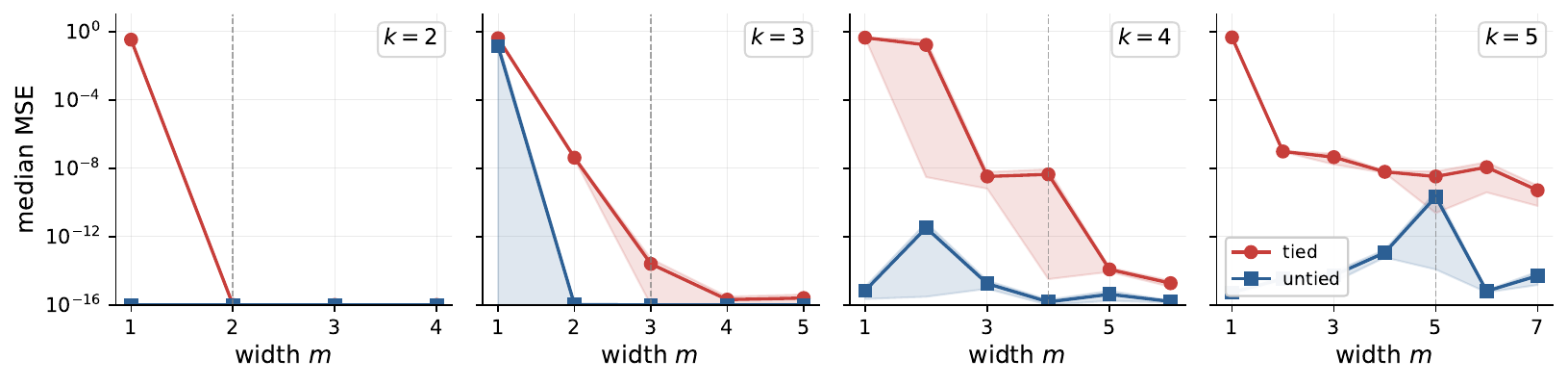}
\caption{Coefficient-space width sweep for \(F_k=x_1^{k-1}x_2\), \(k\in\{2,3,4,5\}\). Untied atoms fit at width \(m=1\). Tied atoms keep a positive loss at \(m=1\); for \(2\le m<k\) their losses are small but nonzero, as expected from border rank two, and they reach machine precision at \(m=k\) for \(k\le3\). Median over two seeds; bands show min--max.}
\label{fig:width_sweep}
\end{figure}

Figure~\ref{fig:width_sweep} shows the predicted separation at width one: untied atoms reach machine precision at \(m=1\) for every degree, while tied atoms keep a clearly positive loss. For \(2\le m<k\), the tied losses are small but nonzero (about \(10^{-7}\) to \(10^{-9}\), with one optimizer outlier at \(k=4\), \(m=2\)). They drop to machine precision at \(m=k\) for \(k\le3\) and at \(m=k+1\) for \(k=4\); for \(k=5\) they stay between about \(10^{-7}\) and \(5\times10^{-10}\) across the tested widths. 
Below its atom rank the target lies outside the tied model class, but for $2 \le m < k$ it lies in the class's closure: the border Waring rank of $x_1^{k-1}x_2$ is two \citep{comas2011rank}, via the limit in Equation~\eqref{eq:border_degeneration} with $\ell=x_1$ and $h=x_2$. The plateau therefore reflects the ill-conditioning of border-rank approximations, whose coefficients must diverge as the residual shrinks, rather than a positive distance to the model class; at $m=1$ the floor is genuine, since the set of symmetric rank-at-most-one tensors is closed.
The untied advantage is specific to single-value-role targets; products with multiple independent value roles require more atoms (Appendix~\ref{app:exp_scope}).

\subsection{Pure powers versus role-asymmetric targets}
\label{sec:exp_pure}

\begin{figure}[t]
\centering
\includegraphics[width=0.8\textwidth]{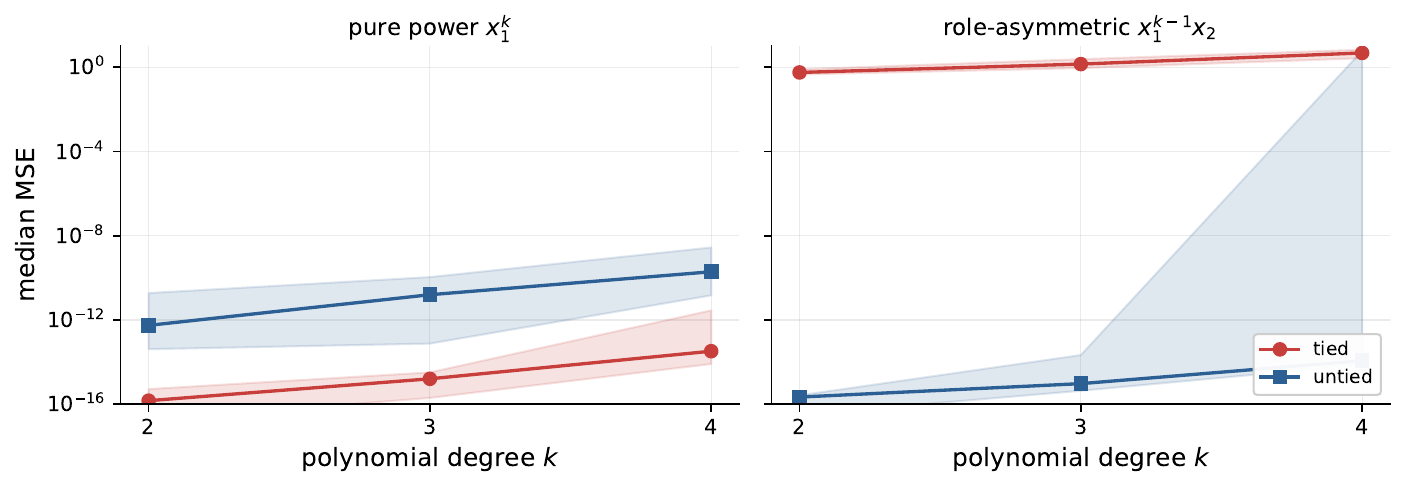}
\caption{Target structure changes the low-width winner. At \(m=1\), tied atoms fit pure powers \(x_1^k\), while untied atoms fit role-asymmetric monomials \(x_1^{k-1}x_2\). Bands show min and max over seeds.}
\label{fig:pure_vs_mixed}
\end{figure}

The previous experiment shows a strong untied advantage on role-asymmetric monomials, but tying is not always worse. At width \(m=1\), we compare tied and untied models on two targets:
\[
    x_1^k
    \qquad\text{and}\qquad
    x_1^{k-1}x_2.
\]
The first is a pure power and is realized exactly by one tied atom; the second is role-asymmetric and is realized exactly by one untied atom. Figure~\ref{fig:pure_vs_mixed} shows the expected reversal: on pure powers, tied models reach machine precision while untied models must align the redundant value direction with the detector; on role-asymmetric targets, untied models fit at width one while tied models cannot. The low-width winner therefore depends on target structure.

\subsection{Separation in gated MLPs}
\label{sec:exp_lift}

The coefficient-space experiments above isolate the atom-rank effect; we now test actual gated MLPs on the same witness \(F_k=x_1^{k-1}x_2\). We compare tied $\mathrm{ReLU}^2$, tied SwiGLU (\(\operatorname{SiLU}(z)z\)), untied ReGLU, and untied SwiGLU, using Gaussian inputs with \(d=4\), \(k\in\{2,3\}\), and widths \(m\in\{1,2,4\}\).

\begin{figure}[t]
\centering
\includegraphics[width=0.8\textwidth]{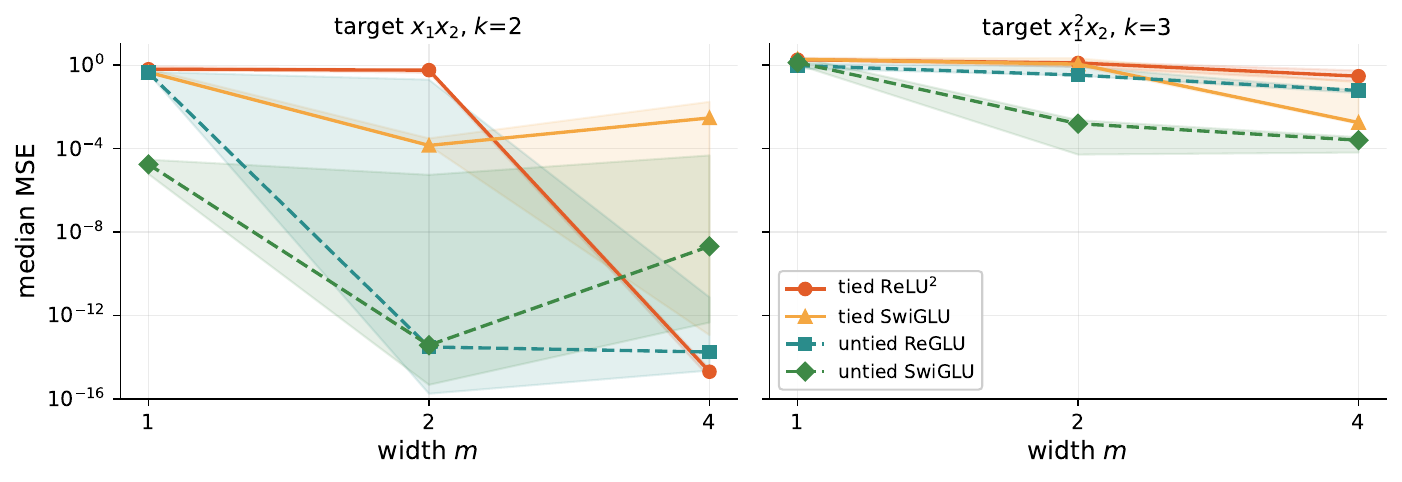}
\caption{Width versus loss for actual gates on \(x_1^{k-1}x_2\). \emph{Left:} for \(k=2\), untied ReGLU and SwiGLU reach near-machine precision at smaller width than tied $\mathrm{ReLU}^2$ and tied SwiGLU. \emph{Right:} for \(k=3\), the exact rank threshold becomes a qualitative ordering; untied SwiGLU is best, and each untied block stays below its same-gate tied counterpart for \(m\ge 2\). Median over three seeds; bands show min--max.}
\label{fig:architectural_lift}
\end{figure}

Figure~\ref{fig:architectural_lift} shows that the exact polynomial threshold becomes a stable qualitative ordering for gated networks. At \(k=2\), untied ReGLU and SwiGLU reach near-machine precision at \(m=2\), and tied $\mathrm{ReLU}^2$ at \(m=4\). These widths are twice the atom ranks \(1\) and \(2\). Both gates satisfy \(g(z)-g(-z)=z\) and \(g(z)z+g(-z)(-z)=z^2\), so each polynomial atom can be built from two gated channels. By the second identity, tied SwiGLU can also represent \(x_1x_2\) exactly at \(m=4\). Its fits here stay near \(10^{-3}\), whereas the same model reaches \(5.5\times10^{-13}\) at width four in the matched control of Appendix~\ref{app:matched}, so this gap is an optimization failure. At \(k=3\), no architecture reaches machine precision. ReLU-gated channels are positively homogeneous of degree two and cannot represent a cubic; a SiLU channel that matches the cubic term also contributes quadratic and higher-degree terms, which finite width cannot cancel exactly. Untied SwiGLU is the best-performing model, and for each gate the untied block lies below its tied counterpart at widths \(m\ge 2\).

A control comparing untied width \(m\) against tied width \(2m\), which matches input-projection parameters, gives a degree-dependent picture: at \(k=2\), tied SwiGLU at width four and untied SwiGLU at width two reach comparable losses, while at \(k=3\) untied SwiGLU at width four is about ten times more accurate than tied SwiGLU at width eight. The figure and numbers are in Appendix~\ref{app:matched}.

%% file: sections/9.conclusions.tex
\section{Conclusion}
\label{sec:conclusion}

We introduced CP-MLPs, a tensor framework for comparing tied and untied multiplicative feed-forward blocks. The framework separates the gate from the choice of whether detector and value directions are shared. Our theory shows that this choice changes width efficiency. Untied atoms can represent indefinite quadratic interactions and role-asymmetric monomials with fewer channels, while tied atoms are naturally matched to pure-power targets. For smooth gates such as SwiGLU, these statements apply degree by degree to Taylor coefficient tensors. Experiments confirm the predicted quadratic width thresholds and the width-one untied fits, show tied losses below the exact monomial rank that are consistent with border rank, and demonstrate that the low-width advantage changes with target structure.

{Selected gated-MLP comparisons retain an untied advantage under input-projection matching, while others attain comparable errors.} These results suggest that the success of GLU-style feed-forward blocks is not only an activation effect: separating detector and value directions changes which interactions are width-efficient. Several open directions remain. Our higher-degree separation is degree-wise rather than a global statement for the full analytic channel, and our experiments use synthetic polynomial targets rather than language-modeling tasks; bridging both gaps, by extending the theory to richer multi-leg atoms and by testing these effects in full Transformer language models, is a natural next step.

%% file: sections/appendix.tex
\appendix

\section{Preliminaries}
This appendix collects the tensor and CP-decomposition notation used
throughout the paper.
\subsection{Tensors and CP decompositions}
\label{app:tensors_cp}

Let $T \in \mathbb{R}^{d_1 \times \cdots \times d_n}$ be an order-$n$ tensor and let
$x_i \in \mathbb{R}^{d_i}$. We write the multilinear contraction of $T$ with
vectors $x_1,\ldots,x_n$ as
\[
T[x_1,\ldots,x_n]
:= T \times_1 x_1 \times_2 \cdots \times_n x_n ,
\]
where $\times_i$ denotes contraction along mode $i$:
\[
\left(T \times_i x_i\right)_{j_1,\ldots,j_{i-1},j_{i+1},\ldots,j_n}
=
\sum_{k=1}^{d_i}
T_{j_1,\ldots,j_{i-1},k,j_{i+1},\ldots,j_n}
(x_i)_k .
\]
Equivalently,
\[
T[x_1,\ldots,x_n]
=
\sum_{j_1=1}^{d_1}\cdots \sum_{j_n=1}^{d_n}
T_{j_1,\ldots,j_n}
(x_1)_{j_1}\cdots (x_n)_{j_n}.
\]
When all modes are contracted with vectors, the result is a scalar. If some modes are left uncontracted, the same formula applies to the remaining modes only; in particular, contracting the $k$ input modes of $T \in \mathbb{R}^{p}\otimes(\mathbb{R}^d)^{\otimes k}$ while leaving the output mode free gives $T[x_1,\ldots,x_k]\in\mathbb{R}^{p}$, the convention used in Section~\ref{sec:degreewise_cp_atoms}. We write $e_i$ for the $i$-th standard basis vector.

A canonical polyadic (CP) decomposition represents a tensor as a sum of rank-one
outer products:
\[
T =
\sum_{a=1}^r
\lambda_a\,
u^{(1)}_a \otimes u^{(2)}_a \otimes \cdots \otimes u^{(n)}_a ,
\]
where $u^{(i)}_a \in \mathbb{R}^{d_i}$ and $\lambda_a \in \mathbb{R}$. Under
contraction, each rank-one term factorizes:
\[
T[x_1,\ldots,x_n]
=
\sum_{a=1}^r
\lambda_a
\prod_{i=1}^n
\langle u^{(i)}_a, x_i\rangle .
\]
Thus a CP decomposition expresses a multilinear map as a sum of separable
products of linear forms.

This is the tensor structure used throughout the paper. For example, a single
CP-MLP channel at degree $k$ contributes a rank-one raw tensor of the form
\[
w_a \otimes u_a^{\otimes(k-1)} \otimes v_a ,
\]
whose diagonal evaluation gives
\[
\bigl(w_a \otimes u_a^{\otimes(k-1)} \otimes v_a\bigr)[x,\ldots,x]
=
w_a (u_a^\top x)^{k-1}(v_a^\top x).
\]
A width-$m$ CP-MLP sums $m$ such channel-wise rank-one contributions.

\paragraph{Symmetrization.}
For any tensor $T \in (\mathbb{R}^d)^{\otimes k}$, define
\[
\Sym(T) = \frac{1}{k!} \sum_{\pi \in S_k} \pi \cdot T.
\]

\paragraph{Symmetric CP rank.}
A symmetric tensor $T$ has symmetric rank $r$ if $r$ is the smallest integer such that
\[
T = \sum_{a=1}^r \lambda_a u_a^{\otimes k}.
\]

\paragraph{Matricization lower bound.}
For any tensor $T$, the rank of any flattening provides a lower bound on its CP rank~\citep{kolda2009tensor}.

\section{Additional framework details}
\subsection{Symmetrization details}
\label{app:cp_mlp_symmetrization}

Let \(S_k\) denote the symmetric group on \(k\) elements. For a permutation
\(\pi\in S_k\) and a tensor \(T\in\mathbb R^p\otimes(\mathbb R^d)^{\otimes k}\),
write \(\pi\cdot T\) for the tensor obtained by permuting the \(k\) input modes
according to \(\pi\), leaving the output mode fixed.

\begin{definition}[Input-symmetric tensor]\label{def:input_symmetric}
A tensor
\[
    T\in\mathbb R^p\otimes(\mathbb R^d)^{\otimes k}
\]
is input-symmetric if \(\pi\cdot T=T\) for every \(\pi\in S_k\). The space of
input-symmetric tensors is denoted
\[
    \mathbb R^p\otimes\Sym^k(\mathbb R^d).
\]
\end{definition}

\begin{definition}[Input symmetrization]
The input symmetrization of
\(T\in\mathbb R^p\otimes(\mathbb R^d)^{\otimes k}\) is
\[
    \Sym(T)
    :=
    \frac{1}{k!}
    \sum_{\pi\in S_k}
    \pi\cdot T.
\]
\end{definition}

\begin{lemma}[Diagonal evaluation sees only the input-symmetric part]
\label{lem:sym_diag}
For every
\[
    T\in\mathbb R^p\otimes(\mathbb R^d)^{\otimes k}
\]
and every \(x\in\mathbb R^d\),
\[
    T[x,\ldots,x]
    =
    \Sym(T)[x,\ldots,x].
\]
\end{lemma}

\begin{proof}
For any \(\pi\in S_k\), all input slots of \(\pi\cdot T\) receive the same
vector \(x\). Hence
\[
    (\pi\cdot T)[x,\ldots,x]
    =
    T[x,\ldots,x].
\]
Averaging over all \(\pi\in S_k\) gives the claim.
\end{proof}

\begin{lemma}[Quadratic diagonal evaluation]
\label{lem:matrix_sym_diag}
For every \(A\in\mathbb R^{d\times d}\) and every \(x\in\mathbb R^d\),
\[
    x^\top A x
    =
    x^\top \frac{A+A^\top}{2}x.
\]
\end{lemma}

\begin{proof}
Let
\[
    S=\frac12(A-A^\top).
\]
Then \(S^\top=-S\). Since \(x^\top Sx\) is a scalar,
\[
    x^\top Sx
    =
    (x^\top Sx)^\top
    =
    x^\top S^\top x
    =
    -x^\top Sx.
\]
Therefore \(x^\top Sx=0\). The result follows by writing
\[
    A=\frac12(A+A^\top)+S.
\]
\end{proof}

\paragraph{Derivative tensors.}
For an analytic scalar function, the degree-\(k\) Taylor coefficient tensor is
equivalently the \(k\)-th derivative tensor up to the factor \(k!\). In
particular, for a homogeneous tensor polynomial
\[
    f_T(x)=T[x,\ldots,x],
\]
we have
\[
    D^k f_T(0)=k!\,\Sym(T).
\]
Thus the derivative tensor also records only the input-symmetric part of the
raw coefficient tensor. This is the differential version of
Lemma~\ref{lem:sym_diag}.

\section{Expressivity proofs}

\subsection{Scalar-output reduction}
\label{app:scalar_projection}

\begin{lemma}[Output projection cannot increase channel rank]
\label{lem:output_projection_lower_bound}
For every \(T\in\mathcal T_k\) and every \(\eta\in(\mathbb R^p)^*\),
\[
    r_{\mathrm{tied}}\bigl((\eta\otimes I)T\bigr)
    \le
    r_{\mathrm{tied}}(T),
    \qquad
    r_{\mathrm{untied}}\bigl((\eta\otimes I)T\bigr)
    \le
    r_{\mathrm{untied}}(T).
\]
\end{lemma}

\begin{proof}
Suppose first that \(T\) has a tied representation with \(m\) channels:
\[
    T
    =
    \sum_{a=1}^{m}
    w_a\otimes u_a^{\otimes k}.
\]
Applying \(\eta\) to the output mode gives
\[
    (\eta\otimes I)T
    =
    \sum_{a=1}^{m}
    \eta(w_a)\,u_a^{\otimes k}.
\]
This is a scalar tied representation with at most \(m\) atoms; atoms with
\(\eta(w_a)=0\) simply drop out.  Minimizing over tied representations of
\(T\) gives the first inequality.

The detector--value case is identical.  If
\[
    T
    =
    \sum_{a=1}^{m}
    w_a\otimes
    \Sym\bigl(u_a^{\otimes(k-1)}\otimes v_a\bigr),
\]
then
\[
    (\eta\otimes I)T
    =
    \sum_{a=1}^{m}
    \eta(w_a)\,
    \Sym\bigl(u_a^{\otimes(k-1)}\otimes v_a\bigr),
\]
which is a scalar detector--value representation with at most \(m\)
atoms.
\end{proof}

For the special embedding of a scalar target into one output coordinate,
the ranks are exactly preserved.

\begin{corollary}[Scalar separations embed into vector targets]
\label{cor:scalar_to_vector}
Let \(S\in\Sym^k(\mathbb R^d)\), and let
\[
    T=e_o\otimes S
    \in
    \mathbb R^p\otimes\Sym^k(\mathbb R^d).
\]
Then
\[
    r_{\mathrm{tied}}(T)
    =
    r_{\mathrm{tied}}(S),
    \qquad
    r_{\mathrm{untied}}(T)
    =
    r_{\mathrm{untied}}(S).
\]
Consequently, any scalar tied--detector--value separation transfers to
the vector-valued CP-MLP setting.
\end{corollary}

\begin{proof}
The lower bounds follow from
Lemma~\ref{lem:output_projection_lower_bound} by taking
\(\eta(y)=y_o\), since
\[
    (\eta\otimes I)T=S.
\]
Thus
\[
    r_{\mathrm{tied}}(S)\le r_{\mathrm{tied}}(T),
    \qquad
    r_{\mathrm{untied}}(S)\le r_{\mathrm{untied}}(T).
\]
For the reverse inequalities, take any scalar representation of \(S\)
and multiply each atom by the output basis vector \(e_o\).  This gives a
vector-valued representation of \(T=e_o\otimes S\) with the same number
of channels.  Hence the ranks are equal.
\end{proof}

\subsection{Bilinear atom-rank proofs}
\label{app:bilinear_width}

\begin{proof}[Proof of Theorem~\ref{thm:bilinear_rank_bounds}]
We prove the tied and untied rank bounds separately.

\paragraph{Tied blocks.}
A tied bilinear block of width \(m\) represents a matrix of the form
\[
    M
    =
    \sum_{a=1}^{m}
    \alpha_a u_a u_a^\top .
\]
Each term \(\alpha_a u_a u_a^\top\) has rank at most one. Therefore
\[
    \operatorname{rank}(M)
    \le
    \sum_{a=1}^{m}
    \operatorname{rank}(\alpha_a u_a u_a^\top)
    \le
    m.
\]
Since \(M\in\mathbb R^{d\times d}\), also
\[
    \operatorname{rank}(M)\le d.
\]
Thus a tied block satisfies
\[
    \operatorname{rank}(M)\le \min(m,d).
\]

This bound is tight. Let \(r=\min(m,d)\). Choosing
\[
    u_a=e_a,
    \qquad
    \alpha_a=1,
    \qquad
    a=1,\ldots,r,
\]
gives
\[
    M
    =
    \sum_{a=1}^{r}
    e_a e_a^\top,
\]
which has rank \(r=\min(m,d)\).

\paragraph{Untied blocks.}
An untied bilinear block of width \(m\) represents a matrix of the form
\[
    M
    =
    \sum_{a=1}^{m}
    M_a,
    \qquad
    M_a
    =
    \frac{\alpha_a}{2}
    \left(
        u_a v_a^\top+v_a u_a^\top
    \right).
\]
For any \(z\in\mathbb R^d\),
\[
    M_a z
    =
    \frac{\alpha_a}{2}
    \left(
        u_a(v_a^\top z)
        +
        v_a(u_a^\top z)
    \right)
    \in
    \operatorname{span}\{u_a,v_a\}.
\]
Therefore
\[
    \operatorname{rank}(M_a)\le 2.
\]
Using subadditivity of rank,
\[
    \operatorname{rank}(M)
    \le
    \sum_{a=1}^{m}
    \operatorname{rank}(M_a)
    \le
    2m.
\]
Since \(M\in\mathbb R^{d\times d}\),
\[
    \operatorname{rank}(M)\le d.
\]
Hence
\[
    \operatorname{rank}(M)\le \min(2m,d).
\]

This bound is tight. Let \(r=\min(2m,d)\).

If \(r=2s\) is even, choose
\[
    u_a=e_{2a-1},
    \qquad
    v_a=e_{2a},
    \qquad
    \alpha_a=1,
    \qquad
    a=1,\ldots,s.
\]
Then
\[
    M
    =
    \frac12
    \sum_{a=1}^{s}
    \left(
        e_{2a-1}e_{2a}^\top
        +
        e_{2a}e_{2a-1}^\top
    \right).
\]
This is a direct sum of \(s\) independent \(2\times2\) off-diagonal
blocks, each of rank two. Therefore \(\operatorname{rank}(M)=2s=r\).

If \(r=2s+1\) is odd, use the same \(s\) off-diagonal blocks and add one
diagonal atom by taking
\[
    u_{s+1}=v_{s+1}=e_{2s+1},
    \qquad
    \alpha_{s+1}=1.
\]
This adds one independent rank-one block, so the resulting matrix has
rank \(2s+1=r\). Thus the untied rank bound is tight.
\end{proof}

\begin{proof}[Proof of Theorem~\ref{thm:exact_bilinear_expressivity}]
We prove the tied and untied atom ranks separately.

\paragraph{Tied atom rank.}
By Theorem~\ref{thm:bilinear_rank_bounds}, any tied representation of
\(M\) with \(m\) atoms satisfies
\[
    \operatorname{rank}(M)\le m.
\]
Therefore
\[
    r_{\mathrm{tied}}(M)
    \ge
    \operatorname{rank}(M).
\]

For the reverse inequality, use the spectral decomposition of the
symmetric matrix \(M\):
\[
    M
    =
    \sum_{i=1}^{n_+(M)}
    \lambda_i q_i q_i^\top
    -
    \sum_{j=1}^{n_-(M)}
    \mu_j p_j p_j^\top,
    \qquad
    \lambda_i,\mu_j>0.
\]
This is a tied representation using one signed rank-one atom for each
nonzero eigendirection. Hence
\[
    r_{\mathrm{tied}}(M)
    \le
    n_+(M)+n_-(M)
    =
    \operatorname{rank}(M).
\]
Combining the two inequalities gives
\[
    r_{\mathrm{tied}}(M)
    =
    \operatorname{rank}(M)
    =
    n_+(M)+n_-(M).
\]

\paragraph{Untied atom rank: lower bound.}
An untied atom satisfies the identity
\[
    \frac{\alpha}{2}
    \left(
        uv^\top+vu^\top
    \right)
    =
    \frac{\alpha}{4}
    (u+v)(u+v)^\top
    -
    \frac{\alpha}{4}
    (u-v)(u-v)^\top .
\]
Therefore any width-\(m\) untied representation can be written as
\[
    M
    =
    AA^\top - BB^\top
\]
for some \(A,B\in\mathbb R^{d\times m}\), after absorbing scalar signs and
magnitudes into the columns of \(A\) and \(B\).

We claim that this implies
\[
    n_+(M)\le m,
    \qquad
    n_-(M)\le m.
\]
Indeed, on \(\ker(A^\top)\),
\[
    x^\top Mx
    =
    x^\top(AA^\top-BB^\top)x
    =
    \|A^\top x\|^2-\|B^\top x\|^2
    =
    -\|B^\top x\|^2
    \le 0.
\]
If $n_+(M) > \operatorname{rank}(A)$, a subspace $P$ of dimension $n_+(M)$ on which $M$ is positive definite would satisfy $\dim(P \cap \ker A^\top) \ge n_+(M) - \operatorname{rank}(A) > 0$, contradicting the inequality above; hence $n_+(M) \le \operatorname{rank}(A) \le m$.
Similarly, on \(\ker(B^\top)\),
\[
    x^\top Mx
    =
    \|A^\top x\|^2
    \ge 0,
\]
so \(M\) cannot be negative definite on a subspace of dimension larger
than \(\operatorname{rank}(B)\le m\). Hence
\[
    n_-(M)\le m.
\]
Therefore every untied representation satisfies
\[
    m
    \ge
    \max\{n_+(M),n_-(M)\}.
\]
So
\[
    r_{\mathrm{untied}}(M)
    \ge
    \max\{n_+(M),n_-(M)\}.
\]

\paragraph{Untied atom rank: upper bound.}
It remains to construct an untied representation with exactly
\(\max\{n_+(M),n_-(M)\}\) atoms. Use the spectral decomposition
\[
    M
    =
    \sum_{i=1}^{n_+(M)}
    \lambda_i q_i q_i^\top
    -
    \sum_{j=1}^{n_-(M)}
    \mu_j p_j p_j^\top,
    \qquad
    \lambda_i,\mu_j>0.
\]

Pair positive and negative eigendirections as much as possible. For a
paired block
\[
    \lambda q q^\top-\mu p p^\top,
    \qquad
    \lambda,\mu>0,
\]
define
\[
    u=\sqrt{\lambda}\,q+\sqrt{\mu}\,p,
    \qquad
    v=\sqrt{\lambda}\,q-\sqrt{\mu}\,p.
\]
Then
\[
    \frac12
    \left(
        uv^\top+vu^\top
    \right)
    =
    \lambda q q^\top-\mu p p^\top.
\]
Thus one untied atom represents one positive and one negative
eigendirection together.

If any positive eigendirections remain unpaired, represent each
\(\lambda q q^\top\) by the tied special case
\[
    u=v=\sqrt{\lambda}\,q.
\]
If any negative eigendirections remain unpaired, represent each
\(-\mu p p^\top\) by
\[
    u=\sqrt{\mu}\,p,
    \qquad
    v=-\sqrt{\mu}\,p.
\]
Therefore the total number of atoms used is
\[
    \max\{n_+(M),n_-(M)\}.
\]
This proves
\[
    r_{\mathrm{untied}}(M)
    \le
    \max\{n_+(M),n_-(M)\}.
\]

Combining the lower and upper bounds gives
\[
    r_{\mathrm{untied}}(M)
    =
    \max\{n_+(M),n_-(M)\}.
\]
\end{proof}

{\subsection{Exact rank, border rank, and bounded approximation}
\label{app:border_rank}

Fix $k\ge2$ and
$\mathsf f\in\{\mathrm{tied},\mathrm{untied}\}$.
All parameters and closures are real.
We use the coefficient-tensor Frobenius norm;
equivalent norms give the same closure. Define
\[
 \mathcal C_{\mathsf f,m}^{(k)}
 =\{T:r_{\mathsf f}(T)\le m\},
 \qquad
 \underline r_{\mathsf f}(T)
 =\min\{m:T\in\overline{\mathcal C_{\mathsf f,m}^{(k)}}\},
\]
and
\[
 e_{\mathsf f,m}(T)
 =\inf_{S\in\mathcal C_{\mathsf f,m}^{(k)}}\|T-S\|_F
 =\operatorname{dist}
   (T,\overline{\mathcal C_{\mathsf f,m}^{(k)}}).
\]
The error is positive for $m<\underline r_{\mathsf f}(T)$,
zero but unattained for
$\underline r_{\mathsf f}(T)\le m<r_{\mathsf f}(T)$,
and zero and attained for $m\ge r_{\mathsf f}(T)$.
A positive infimum alone does not guarantee a best
approximation~\citep{desilva2008tensor}.

\paragraph{Coefficient sizes.}
For output $w_a\in\mathbb R^p$, define
\[
 A_a^t=\|w_a\|\,\|u_a\|^k,
 \qquad
 A_a^u=\|w_a\|\,\|u_a\|^{k-1}\|v_a\|.
\]
These sizes are invariant under factor rescalings that preserve
the raw atom; for scalar output, $\|w_a\|=|w_a|$.
Let $\mathcal C_{\mathsf f,m}^{(k)}(B)$ contain tensors with
a representation using at most $m$ atoms and total size
$\sum_a A_a^{\mathsf f}\le B$.

\begin{proposition}[Bounded compactness and forced size divergence]
\label{prop:bounded_compactness}
For finite $B\ge0$, both classes
$\mathcal C_{\mathsf f,m}^{(k)}(B)$ are compact. Thus
\[
 \delta_{\mathsf f,m}(B;T)
 =\min_{S\in\mathcal C_{\mathsf f,m}^{(k)}(B)}
   \|T-S\|_F
\]
exists, is positive when $m<r_{\mathsf f}(T)$, and converges
to $e_{\mathsf f,m}(T)$ as $B\to\infty$.
Every fixed-width sequence converging to a target outside
the exact class has $A_{\max}\to\infty$.
\end{proposition}

\begin{proof}
Normalize the input directions and absorb their norms into
the output factors, padding zero atoms as needed.
The tied parameter set becomes
\[
 \{(\widetilde w_1,\ldots,\widetilde w_m):
   \textstyle\sum_a\|\widetilde w_a\|\le B\}
 \times(\mathbb S^{d-1})^m.
\]
It is compact, as is its image under the continuous tensor-sum
map. The untied case uses two unit directions per atom.
Compactness gives the positive minimum outside the class.
The budgeted classes increase to the unrestricted class,
so their distances decrease to $e_{\mathsf f,m}(T)$.
If $A_{\max}$ did not tend to infinity, a subsequence would
have total size at most $mC$ for some finite $C$;
compactness would place its limit in the exact class,
a contradiction.
\end{proof}

\begin{proposition}[Separation under a common coefficient-size budget]
\label{prop:matched_budget}
If $r_{\mathrm{untied}}(T)\le m<r_{\mathrm{tied}}(T)$,
there is a finite $B_0$ such that every $B\ge B_0$ permits
exact untied width-$m$ representation, whereas
$\delta_{\mathrm{tied},m}(B;T)>0$.
As $B\to\infty$, this tied distance has a positive limit
if $m<\underline r_{\mathrm{tied}}(T)$ and tends to zero
otherwise.
\end{proposition}

\begin{proof}
Take $B_0$ to be the total size of an exact untied representation.
Apply Proposition~\ref{prop:bounded_compactness} and the
definition of border rank to the tied class.
\end{proof}

\paragraph{The monomial limit.}
For independent linear forms $\ell,h$,
\begin{equation}
 P_\varepsilon
 =\frac{(\ell+\varepsilon h)^k-(\ell-\varepsilon h)^k}
        {2k\varepsilon}
 \longrightarrow \ell^{k-1}h.
 \label{eq:border_degeneration}
\end{equation}
Hence the tied border rank is at most two.
It is at least two because the first-derivative span of
$\ell^{k-1}h$ has dimension two, while a pure power has
dimension at most one, and this matrix-rank bound survives
closure. Thus
$\underline r_{\mathrm{tied}}(F_k)=2$~\citep{comas2011rank}.
For $k\ge3$, binomial expansion gives error
$O(\varepsilon^2)$ and
$A_{\max}=\Theta(|\varepsilon|^{-1})$,
hence squared error $O(A_{\max}^{-4})$ along this path.
This gives an upper bound; optimality is not established.
For $k=2$, the identity is exact for nonzero $\varepsilon$.

\par}

{\subsection{Closed quadratic classes}
\label{app:closed_quadratic}

\begin{proposition}[Scalar quadratic closedness]
\label{prop:quadratic_closed}
For scalar degree-two targets and every width $m$, both tied
and untied classes are closed. Their exact and real border
ranks coincide.
\end{proposition}

\begin{proof}
By Theorem~\ref{thm:exact_bilinear_expressivity}, the tied class
consists of symmetric matrices of rank at most $m$, a closed
condition defined by vanishing $(m+1)$-minors.
The untied class is
\[
 \{M=M^\top:n_+(M)\le m,\ n_-(M)\le m\},
\]
which is closed by continuity of ordered eigenvalues.
Distance from a target outside either class is positive and
attained: a minimizing sequence can be restricted to a bounded
ball, whose intersection with the class is compact.
\end{proof}

For $x\sim\mathcal N(0,I_d)$ and symmetric $H$,
\[
 \mathbb E[(x^\top Hx)^2]
 =2\|H\|_F^2+(\operatorname{tr}H)^2.
\]
A positive coefficient-norm gap therefore implies a positive
optimal population MSE. An observed finite-sample residual
can also reflect sampling and optimization.
For $Q=\operatorname{diag}(1,1,-1,-1)$, tied rank-$m$ spectral
truncation gives relative squared Frobenius error $(4-m)/4$
for $0\le m\le4$; the untied error is $1/2$ at $m=1$
and zero at $m=2$.

\par}

{\subsection{Universal comparison and independent interaction blocks}
\label{app:block_family}

\begin{proposition}[Universal exact- and border-rank comparison]
For every degree-$k$ coefficient tensor $T$, with $k\ge2$,
\[
 r_{\mathrm{untied}}(T)
 \le r_{\mathrm{tied}}(T)
 \le k\,r_{\mathrm{untied}}(T),
 \qquad
 \underline r_{\mathrm{untied}}(T)
 \le \underline r_{\mathrm{tied}}(T)
 \le 2\,\underline r_{\mathrm{untied}}(T).
\]
Both upper constants are attained by $F_k$.
\end{proposition}

\begin{proof}
Containment gives the lower bounds, including after closure.
For independent $u,v$, a change of coordinates maps
$(u^\top x)^{k-1}(v^\top x)$ to $F_k$, which has a real
$k$-power decomposition by
Theorem~\ref{thm:monomial_separation}; collinear directions
require at most one power.
Applying these decompositions to each atom, with its output
vector, proves the exact upper bound.
Equation~\eqref{eq:border_degeneration} puts each untied atom
in the closure of the tied width-two class.
Summing and taking closure gives the border upper bound.
The ranks of $F_k$ establish sharpness.
\end{proof}

\begin{theorem}[Independent interaction blocks]
\label{thm:block_family}
Let $k\ge2$, $s\ge1$, $d\ge2s$, and
$c_i\in\mathbb R\setminus\{0\}$.
For linearly independent real forms
$\ell_1,h_1,\ldots,\ell_s,h_s$, define
\[
 F_{k,s}(x)=\sum_{i=1}^s c_i\ell_i(x)^{k-1}h_i(x).
\]
Then
\[
 \bigl(
 r_{\mathrm{untied}},
 r_{\mathrm{tied}},
 \underline r_{\mathrm{untied}},
 \underline r_{\mathrm{tied}}
 \bigr)(F_{k,s})
 =(s,ks,s,2s).
\]
\end{theorem}

\begin{proof}
Use the $2s$ independent forms as coordinates.
The first derivatives span the $2s$ distinct monomials
$x_{2i-1}^{k-2}x_{2i}$ and $x_{2i-1}^{k-1}$,
including when $k=2$.
Each tied atom contributes at most one derivative direction
and each untied atom at most two.
Since these matrix-rank bounds survive closure,
\[
 \underline r_{\mathrm{tied}}\ge2s,
 \qquad
 \underline r_{\mathrm{untied}}\ge s.
\]
One untied atom per summand gives
$r_{\mathrm{untied}}\le s$;
Equation~\eqref{eq:border_degeneration} gives
$\underline r_{\mathrm{tied}}\le2s$;
and real $k$-power decompositions give
$r_{\mathrm{tied}}\le ks$.

For $k\ge3$, absorb $c_i$ into $x_{2i}$ and apply the
additivity theorem for pairwise coprime
monomials~\citep[Theorem~3.2]{carlini2012coprime}:
the complex Waring rank is $ks$, so the real rank is
at least $ks$.
Extra coordinates cannot lower rank, since restricting them
to zero preserves a decomposition.
For $k=2$, each block has one positive and one negative
eigenvalue, and Theorem~\ref{thm:exact_bilinear_expressivity}
gives tied rank $2s=ks$.
\end{proof}

Thus the tied approximation gap is positive at $s\le m<2s$,
while the untied family is exact.
At $2s\le m<ks$, the tied infimum is zero but unattained.

\par}

{\subsection{Necessary width bounds for analytic channels}
\label{app:analytic_width}

\begin{proposition}[Degree-wise necessity]
Let $g$ be analytic near zero and let $F$ be represented near
zero by a bias-free width-$m$ CP-MLP in family $\mathsf f$.
Then $F(0)=0$, inactive Taylor components vanish, and
\[
 m\ge\sup_{k\in\mathcal S(g)}
 r_{\mathsf f}(F^{(k)}).
\]
\end{proposition}

\begin{proof}
Equality near zero implies equality of Taylor coefficients.
Each active degree is a sum of at most $m$ atoms, with its
nonzero gate coefficient absorbed into the output factors;
inactive components vanish.
These conditions are necessary, but parameters shared across
degrees prevent choosing the decompositions independently.
\end{proof}

\begin{proposition}[Degree-wise rank one with SiLU-channel width two]
\label{prop:silu_invariant}
The function
\[
 G(x_1,x_2)
 =\bigl[\operatorname{SiLU}(x_1)
       +\operatorname{SiLU}(2x_1)\bigr]x_2
\]
has untied atom rank one at every active Taylor degree,
but its exact untied SiLU-channel width is two.
\end{proposition}

\begin{proof}
The displayed representation uses two channels, and each active
component is a nonzero multiple of $x_1^{k-1}x_2$.
To exclude width one, expand
\[
 \operatorname{SiLU}(z)
 =\tfrac12z+\tfrac14z^2-\tfrac1{48}z^4+O(z^6).
\]
For $A\operatorname{SiLU}(s x_1)x_2$, the coefficients
$b_2,b_3,b_5$ of $x_1x_2,x_1^2x_2,x_1^4x_2$ satisfy
\[
 b_2=As/2,\qquad
 b_3=As^2/4,\qquad
 b_5=-As^4/48,
 \qquad
 \frac{b_2^2b_5}{b_3^3}=-\frac13
\]
when $b_3\ne0$.
For $G$, they are $3/2,5/4,-17/48$, giving
$-51/125\ne-1/3$.

Any width-one representation matching the quadratic term
must have detector and value forms proportional to $x_1,x_2$
in some order, by unique factorization.
The cubic term fixes the detector along $x_1$ and the value
along $x_2$, so the invariant excludes all width-one
representations.
\end{proof}

The obstruction is the gate's fixed coefficient ratios across
degrees. Small held-out function error alone does not bound
every Taylor coefficient.

\par}

\section{Experimental details}
\label{app:exp}

This section documents the experimental methodology in sufficient detail for an independent re-implementation. Code that reproduces every figure and table is available at \url{https://github.com/rarefin/CP-MLPs}.
\subsection{Computational requirements}
\label{app:compute}

All experiments run on a standard CPU; no GPU or specialized hardware is required. The experiments of Section~\ref{sec:exp} and Appendices~\ref{app:exp_scope}--\ref{app:matched} each complete in about one minute, about five minutes in total on a single CPU core. The additional experiments of Appendices~\ref{app:extra_diagnostics}--\ref{app:extra_adam} take about two to three hours in total on one core. Memory usage stays below 1\,GB throughout.

The experiments use standard scientific Python: \texttt{numpy} for linear algebra, \texttt{scipy.optimize.minimize} (L-BFGS-B method) for nonlinear least-squares fitting, and \texttt{matplotlib} for figure generation. The experiments of Appendices~\ref{app:extra_recovery}, \ref{app:extra_random}, and~\ref{app:extra_adam} also use PyTorch on the CPU, for automatic differentiation and for Adam.

\subsection{Data generation}
\label{app:data}

The experiments fit two kinds of targets: \emph{polynomial coefficient tensors} (in coefficient space) and \emph{scalar functions of Gaussian inputs} (in function space). We describe both.

\paragraph{Coefficient-space targets.}
A degree-\(k\) polynomial in \(d\) variables corresponds to a coefficient vector \(T\in\mathbb R^D\) in the basis of degree-\(k\) monomials, where \(D=\binom{d+k-1}{k}\). For \(d=2\) we use the basis indexed by the exponent of \(x_2\), so index \(j\) corresponds to the monomial \(x_1^{k-j}x_2^j\). The role-asymmetric witness \(F_k=x_1^{k-1}x_2\) of Theorem~\ref{thm:monomial_separation} therefore has a single nonzero entry at index \(j=1\); the pure-power target \(x_1^k\) has its single nonzero entry at \(j=0\).

The model's prediction in coefficient space is the coefficient vector of its parameterized polynomial, computed analytically from the architecture's atom form (Section~\ref{app:arch_coef}). The loss is the squared \(\ell_2\) distance between the predicted coefficient vector and the target.

\paragraph{Function-space targets.}
For experiments using actual Gaussian inputs (Sections~\ref{sec:exp_pure} and~\ref{sec:exp_lift}), we sample inputs \(X\in\mathbb R^{N\times d}\) with each row drawn i.i.d.\ from \(\mathcal N(0,I_d)\). The sample size \(N\) is given in Table~\ref{tab:hyperparams}. Targets are evaluated pointwise: for \(y(x)=x_1^{k-1}x_2\), we set \(y_i=X_{i,1}^{k-1}X_{i,2}\); for the indefinite quadratic experiment, \(y_i=X_i^\top Q X_i\) with \(Q=\operatorname{diag}(\mathbf 1^{n_+},-\mathbf 1^{n_-},\mathbf 0^{d-n_+-n_-})\). The loss is the mean squared error \(\frac{1}{N}\sum_i(\hat y_i-y_i)^2\).

\paragraph{Random seeds.}
Each experiment uses random seeds that are deterministic functions of configuration values. For example, the quadratic experiment uses $\verb|seed = s * 137 + n_plus * 17 + n_minus|$ for seed index \(s\) and signature \((n_+,n_-)\); the width-sweep experiment uses $\verb|seed = s * 37 + k * 11|$, shared across the widths of a sweep. Fixed seeds make every run reproducible with the same library versions.

\subsection{Model implementations}
\label{app:archs}

Each model is parameterized by a flat parameter vector \(\theta\in\mathbb R^P\) that the optimizer treats as the only variable. The total parameter count \(P\) depends on the architecture and the width \(m\).

\paragraph{Function-space architectures.}
Given input \(X\in\mathbb R^{N\times d}\) and width \(m\), all architectures follow the CP-MLP structure of Definition~\ref{def:cp_mlp}: project inputs through one or two linear maps, apply an elementwise nonlinearity, multiply per channel, and combine via scalar output coefficients \(c\in\mathbb R^m\) (the scalar specialization of the output direction \(w_a\)).

For \textbf{tied} polynomial atoms with detector matrix \(U\in\mathbb R^{m\times d}\) and coefficients \(c\):
\[
    \hat y_i = \sum_{a=1}^m c_a\,(u_a^\top x_i)^k,
    \qquad
    \theta=(\operatorname{vec}(U),c)\in\mathbb R^{m(d+1)}.
\]

For \textbf{untied} polynomial atoms with detector \(U\), value \(V\in\mathbb R^{m\times d}\), and coefficients \(c\):
\[
    \hat y_i = \sum_{a=1}^m c_a\,(u_a^\top x_i)^{k-1}(v_a^\top x_i),
    \qquad
    \theta=(\operatorname{vec}(U),\operatorname{vec}(V),c)\in\mathbb R^{m(2d+1)}.
\]

For the four gated architectures in Section~\ref{sec:exp_lift}, replace the polynomial atoms with their gated analogues from Table~\ref{tab:cp_mlp_variants}:
{\small
\begin{align*}
\text{tied $\mathrm{ReLU}^2$:}\quad
&\hat y=\sum_a c_a\,\mathrm{ReLU}(u_a^\top x)^2,
&
\text{untied ReGLU:}\quad
&\hat y=\sum_a c_a\,\mathrm{ReLU}(u_a^\top x)(v_a^\top x),
\\
\text{tied SwiGLU:}\quad
&\hat y=\sum_a c_a\,\mathrm{SiLU}(u_a^\top x)(u_a^\top x),
&
\text{untied SwiGLU:}\quad
&\hat y=\sum_a c_a\,\mathrm{SiLU}(u_a^\top x)(v_a^\top x).
\end{align*}
}
with \(\mathrm{SiLU}(z)=z\sigma(z)\) and \(\sigma(z)=1/(1+e^{-z})\). Tied gated architectures have parameter count \(m(d+1)\); untied gated architectures have \(m(2d+1)\). The matched-parameter control of Section~\ref{sec:exp_lift} pairs untied width \(m\) against tied width \(2m\), equalizing input-projection parameters at \(2md\).

\paragraph{Coefficient-space implementations.}
\label{app:arch_coef}
For coefficient-space experiments we compute the coefficient vector of each parameterized polynomial directly, avoiding sampling. For tied atoms, the multinomial theorem gives
\[
    (u^\top x)^k=\sum_{|\alpha|=k}\binom{k}{\alpha}u^\alpha x^\alpha,
    \qquad
    \binom{k}{\alpha}=\frac{k!}{\alpha_1!\cdots\alpha_d!}.
\]
For untied detector-value atoms,
\[
    (u^\top x)^{k-1}(v^\top x)
    =
    \left(\sum_{|\beta|=k-1}\binom{k-1}{\beta}u^\beta x^\beta\right)\sum_{j=1}^d v_j x_j,
\]
which expands into a degree-\(k\) multi-index sum. The total predicted coefficient vector is \(\sum_a c_a\) times the per-atom coefficient vector, and the loss is the squared \(\ell_2\) distance from the target.

\subsection{Optimization protocol}
\label{app:opt}

The experiments of Section~\ref{sec:exp} and Appendices~\ref{app:exp_scope}--\ref{app:matched} minimize the squared loss using \texttt{scipy.optimize.minimize} with method \texttt{L-BFGS-B}. The iteration limit and tolerances are: indefinite quadratics \(3000\), \(\texttt{ftol}=10^{-22}\), \(\texttt{gtol}=10^{-15}\); width sweep \(2000\), \(10^{-22}\), \(10^{-18}\); pure-vs-mixed \(5000\), \(10^{-20}\), \(10^{-15}\); gated MLPs \(1500\), \(10^{-18}\), \(10^{-12}\); matched control \(1500\), \(10^{-20}\), \(10^{-12}\); scope boundary \(800\), \(10^{-22}\), \(10^{-15}\). Appendices~\ref{app:extra_diagnostics}--\ref{app:extra_adam} describe their own settings. For each random seed, we run multiple restarts with parameter initialization \(\theta_0\sim\mathcal N(0,0.25)\) (applied independently to each entry of \(\theta\)). We record the best-of-restarts loss per seed and aggregate (median, min, max) across seeds.

\subsection{Per-experiment configurations}
\label{app:hyper}

\begin{table}[h]
\centering
\caption{Hyperparameters for each experiment.}
\label{tab:hyperparams}
\small
\begin{tabular}{@{}lccccc@{}}
\toprule
Experiment & Section & Input dim \(d\) & Sample size & Seeds & Restarts \\
\midrule
Indefinite quadratic targets   & \S\ref{sec:exp_quad}    & \(8\) & \(256\)       & \(3\)  & \(8\)  \\
Role-asymmetric monomial targets & \S\ref{sec:exp_waring}  & \(2\) & coef.\ space  & \(2\)  & \(5\) \\
Pure-vs-mixed reversal         & \S\ref{sec:exp_pure}    & \(4\) & \(256\)       & \(10\) & \(5\)  \\
Gated MLPs             & \S\ref{sec:exp_lift}    & \(4\) & \(200\)       & \(3\)  & \(3\)  \\
Matched-parameter control (Gated MLPs)      & \S\ref{sec:exp_lift}    & \(4\) & \(200\)       & \(3\)  & \(3\)  \\
Scope boundary            & \S\ref{app:exp_scope}  & \(3\) & coef.\ space  & \(2\)  & \(4\)\(^{*}\) \\
\bottomrule
\end{tabular}
\par\smallskip
{\footnotesize \(^{*}\)Twelve restarts for the untied model at \(m=1\).}
\end{table}

For coefficient-space experiments, fits operate directly on the \(\binom{d+k-1}{k}\) monomial coefficients of the target polynomial, eliminating sampling and function-evaluation noise. For experiments with finite samples, inputs are drawn from \(\mathcal N(0,I_d)\) once per seed and held fixed across restarts within that seed.

\subsection{Convergence criterion}
\label{app:conv}

We report best-of-restarts MSE without applying a separate convergence threshold. A configuration is described as \emph{fitting} the target if the residual reaches machine precision (\(\mathrm{MSE}<10^{-8}\)) and as \emph{failing} if the residual stays above \(10^{-1}\). The intermediate regime is interpreted as a partial fit due to optimizer-landscape difficulty rather than a representational limitation, and is discussed for the \(k=5\) case in Section~\ref{sec:exp_waring}.

To distinguish the plateau of Section~\ref{sec:exp_waring} from a slow-optimization artifact, we re-ran the failed configurations with \(5\times\) the L-BFGS iteration budget. The residual stayed within \(10\%\) of its plateau value across all configurations tested, confirming that the plateau is robust to optimization budget.

\subsection{Scope of the role-asymmetric separation}
\label{app:exp_scope}
Theorem~\ref{thm:monomial_separation} shows that one untied detector-value atom suffices for the role-asymmetric witness \(F_{dv} = x_1^{k-1}x_2\), where one direction acts as a repeated detector and one direction supplies the single value. This appendix tests the boundary of that statement: targets with multiple independent value roles do \emph{not} reduce to a single untied atom.

Concretely, at \(k=3\) we compare tied, untied, and a multi-leg oracle (the smallest atom family rich enough to express any product of \(k\) linear forms) on the role-asymmetric witness \(F_{dv}\) and on the square-free product
\[
    F_{sf} = x_1 x_2 \cdots x_k,
\]
in which each of the \(k\) variables appears exactly once.

\begin{figure}[t]
\centering
\includegraphics[width=\textwidth]{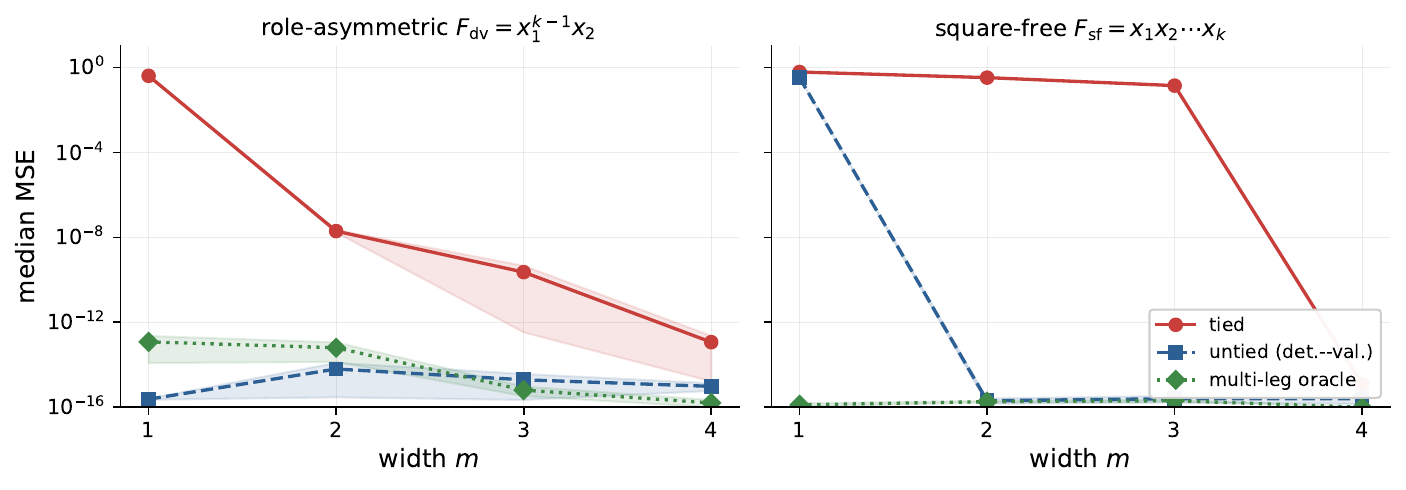}
\caption{Scope boundary at \(k=3\). Untied atoms are rank one on the role-asymmetric witness \(F_{dv} = x_1^{k-1}x_2\) but not on the square-free product \(F_{sf} = x_1x_2\cdots x_k\); the multi-leg oracle is rank one on both.}
\label{fig:scope}
\end{figure}

Figure~\ref{fig:scope} and Table~\ref{tab:scope} confirm three regimes. On \(F_{dv}\), untied atoms and the multi-leg oracle both fit at \(m=1\). Tied atoms keep a positive loss at \(m=1\), reach small but nonzero losses at \(m=2\) and \(m=3\) (about \(10^{-8}\) and \(10^{-10}\), consistent with border rank two), and reach about \(10^{-13}\) at \(m=4\). On \(F_{sf}\), the multi-leg oracle fits at \(m=1\), untied requires \(m=2\), and tied requires \(m\ge 2^{k-1}\), matching its Waring rank (by Proposition~\ref{prop:waring_rank_monomials}). The untied advantage is therefore specialized to targets with a single value role and does not extend to arbitrary products.

\begin{table}[t]
\centering
\caption{Scope boundary at \(k=3\). Median best-of-restarts MSE over two seeds.}
\label{tab:scope}
\small
\input{figures/expressivity2/table_scope}
\end{table}

\subsection{
{Matched input-projection control}
in gated MLPs}
\label{app:matched}

\begin{figure}[t]
\centering
\includegraphics[width=\textwidth]{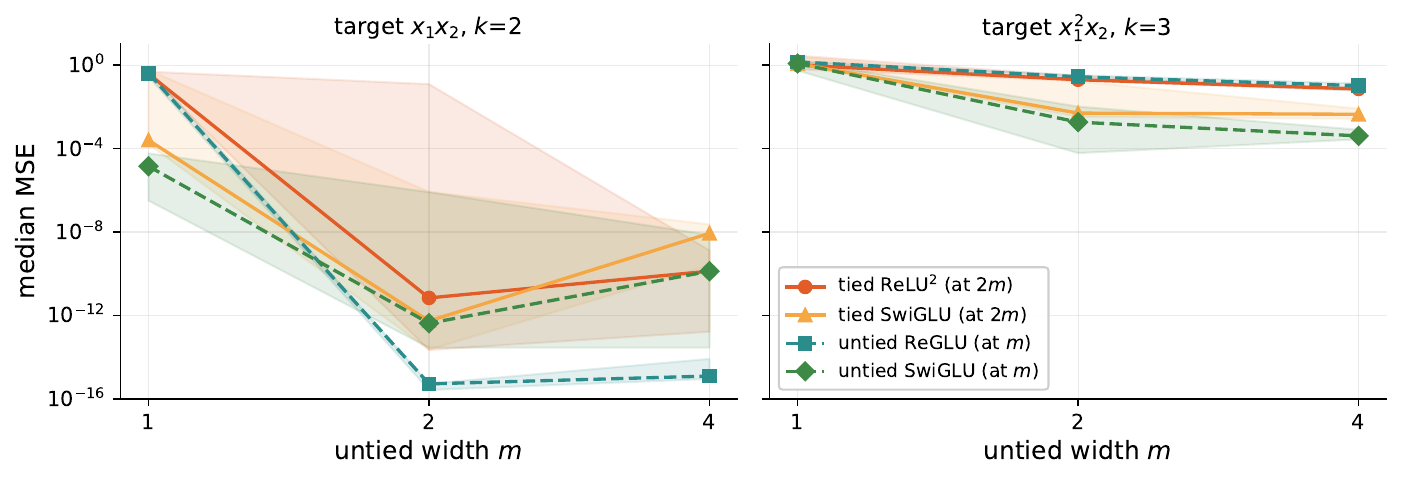}
\caption{
{Matched input-projection control on
\(x_1^{k-1}x_2\): untied width \(m\) (dashed) versus tied width
\(2m\) (solid). Tied SwiGLU uses \(\operatorname{SiLU}(z)z\).
Median over three seeds; bands show min--max.
Total parameter counts are \(m(2d+1)\) and \(2m(d+1)\),
respectively.}
}
\label{fig:matched}
\end{figure}

{Untied architectures use two input projections per channel.
We compare untied width \(m\) with tied width \(2m\),
matching input-projection counts at \(2md\).
The tied model has \(m\) additional scalar-output parameters.
\par}

{Figure~\ref{fig:matched} shows that the ordering
depends on degree and width.
At \(k=2\), tied SwiGLU at width four and untied SwiGLU
at width two attain comparable median losses,
\(5.5\times10^{-13}\) and \(4.3\times10^{-13}\), respectively.
At \(k=3\), untied width four reaches \(4.0\times10^{-4}\),
compared with \(4.4\times10^{-3}\) for tied width eight,
an approximately \(10.8\)-fold difference.
These fixed-budget fits show a selected untied advantage,
not a uniform separation or minimum-width certificate.
\par}

{\subsection{Border-rank diagnostics}
\label{app:extra_diagnostics}

The additional coefficient-space experiments use the apolar norm.
If $f(x)=\sum_{|\alpha|=k}f_\alpha x^\alpha$, then
\[
 \|f\|_{\mathrm{ap}}^2
 =\sum_{|\alpha|=k}\frac{\alpha!}{k!}f_\alpha^2.
\]
This is the Frobenius norm of the symmetric coefficient tensor
and satisfies $\|(u^\top x)^k\|_{\mathrm{ap}}=\|u\|^k$.
Reported losses are relative squared errors
$\|\widehat f-f\|_{\mathrm{ap}}^2/\|f\|_{\mathrm{ap}}^2$.
Unlike the unweighted monomial-coordinate norm used in
Section~\ref{sec:exp_waring}, this convention accounts for
multinomial multiplicities. The two norms have the same closure,
but numerical losses need not agree.

For the size diagnostic, each configuration has 27 candidate
fits: 12 unregularized initializations and five at each of three
penalties on $\sum_a A_a^2$.
We record the smallest loss and the smallest $A_{\max}$ among
candidates with loss at most $1.02$ times that minimum.
At the selected near-best candidate, we also report the intrinsic
condition number $\kappa=1/\sigma_{\min}(\mathcal T)$ for the
normalized Terracini/Jacobian matrix after removing scaling
directions~\citep{vannieuwenhoven2017condition}.
A small singular value indicates sensitivity of decomposition
parameters; it is not a certificate of global optimality.
The notation ``n.i.'' means that local nonuniqueness makes this
diagnostic uninformative.

\begin{table}[t]
\centering\small
\caption{Coefficient-size diagnostics for $F_3=x_1^2x_2$.
Exact and border ranks determine the regimes; losses and sizes
describe the candidate fits.}
\label{tab:border_single}
\setlength{\tabcolsep}{4pt}
\begin{tabular}{@{}lclrrrr@{}}
\toprule
Family & $m$ & Regime & Best loss & Best $A_{\max}$
& Near-best $A_{\max}$ & $\kappa$ \\
\midrule
Untied & 1 & Exact control
& $1.6\!\times\!10^{-29}$ & 1.00 & 1.00 & 1.7 \\
Tied & 1 & Positive infimum
& 0.5556 & 0.385 & 0.385 & 1.5 \\
Tied & 2 & Zero, unattained
& $5.4\!\times\!10^{-12}$ & 83.17 & 83.17
& $4.4\!\times\!10^7$ \\
Tied & 3 & Exact width
& $1.5\!\times\!10^{-28}$ & 0.471 & 0.471 & n.i. \\
\bottomrule
\end{tabular}
\end{table}

Table~\ref{tab:border_single} follows the three regimes for the
original monomial.
At tied width one, the positive floor is attained because the
class is closed.
At width two, the smallest near-best candidate has large size;
this is consistent with the forced-divergence statement for any
sequence whose error tends to zero.
A finite collection of large candidates does not exclude a
smaller-size candidate that the optimizer missed.
At exact width three, a bounded near-zero fit is recovered.

\begin{table}[t]
\centering\small
\caption{The same diagnostic for
$F_{3,2}=x_1^2x_2+x_3^2x_4$.
Its tied border width is four and tied exact width is six.}
\label{tab:border_blocks}
\setlength{\tabcolsep}{4pt}
\begin{tabular}{@{}lclrrrr@{}}
\toprule
Family & $m$ & Regime & Best loss & Best $A_{\max}$
& Near-best $A_{\max}$ & $\kappa$ \\
\midrule
Untied & 2 & Exact control
& $6.4\!\times\!10^{-19}$ & 1.00 & 1.00 & 1.7 \\
Tied & 2 & Positive infimum
& 0.5000 & 46.95 & 2.27 & $8.6\!\times\!10^2$ \\
Tied & 3 & Positive infimum
& 0.2778 & 66.30 & 2.27 & $8.6\!\times\!10^2$ \\
Tied & 4 & Zero, unattained
& $1.1\!\times\!10^{-10}$ & 47.74 & 47.736
& $8.3\!\times\!10^6$ \\
Tied & 5 & Zero, unattained
& $9.3\!\times\!10^{-11}$ & 34.40 & 34.405
& $1.9\!\times\!10^{13}$ \\
Tied & 6 & Exact width
& $8.3\!\times\!10^{-16}$ & 3.72 & 3.72 & n.i. \\
\bottomrule
\end{tabular}
\end{table}

For the independent-block target, Table~\ref{tab:border_blocks}
places the positive-to-zero infimum transition at width four,
as predicted by Theorem~\ref{thm:block_family}.
The large best-fit size at width two does not contradict the
positive gap: this candidate approximates one block through
a degeneration while leaving the other unfitted, at loss $0.5$.
Its condition number is $7.9\times10^6$, whereas a candidate
within $2\%$ of the same loss has size $2.27$.
The theorem identifies the regime; the diagnostics describe
how individual fits approach it.

As a closed-class control, tied fits to
$Q=\operatorname{diag}(1,1,-1,-1)$ yield relative coefficient
errors $0.7500$, $0.5000$, and $0.2500$ at widths one, two,
and three, matching $(4-m)/4$.
Penalty sweeps along the monomial degeneration give log--log
slopes $-4.003$, $-4.002$, and $-4.002$ for squared error
versus $A_{\max}$ at $k=3,4,5$, consistent with the constructive
$O(A_{\max}^{-4})$ path in
Equation~\eqref{eq:border_degeneration}.
These slopes do not prove an optimal trade-off.

\par}

{\subsection{Recovery of known representations}
\label{app:extra_recovery}

To separate optimization from expressivity, we retain generating
parameters for targets constructed from three atoms.
These witnesses establish width-three representability,
not minimal rank three.
Fitting uses float64 L-BFGS-B with automatic differentiation,
$d=3$, and 60 restarts per configuration.
Recovery means relative apolar loss below $10^{-10}$.
Table~\ref{tab:recovery_rates} reports recovery fractions for
the respective target constructions in each family;
it should not be read as a comparison on identical targets.

\begin{table}[t]
\centering\small
\caption{Recovery fractions for configurations with a width-three
generating witness. Witness losses are approximately $10^{-32}$.
Widths four and five pad the representation with additional channels.}
\label{tab:recovery_rates}
\begin{tabular}{@{}c|ccc|ccc|r@{}}
\toprule
& \multicolumn{3}{c|}{Tied width}
& \multicolumn{3}{c|}{Untied width}
& Best untied loss \\
$k$ & 3 & 4 & 5 & 3 & 4 & 5 & at width 3 \\
\midrule
3 & 0.33 & 0.32 & 0.70 & 0.80 & 0.92 & 1.00
& $1.2\times10^{-20}$ \\
4 & 0.42 & 0.67 & 0.75 & 0.17 & 0.65 & 0.88
& $2.4\times10^{-18}$ \\
5 & 0.77 & 0.80 & 0.85 & 0.00 & 0.02 & 0.00
& $1.7\times10^{-5}$ \\
6 & 0.08 & 0.20 & 0.22 & 0.00 & 0.00 & 0.00
& $8.1\times10^{-9}$ \\
\bottomrule
\end{tabular}
\end{table}

No untied width-three restart reaches the tolerance at $k=5$
or $k=6$, despite the generating witnesses.
At $k=5$, widening yields one recovery in 60 attempts at
width four and none at width five.
Thus the tested extra widths do not reliably resolve the
optimization difficulty.
In a separate structured example, $F_{4,2}$ has exact untied
rank two, but eight restarts all return loss $0.25$.
A 40-restart run recovers the representation ten times,
with best loss $5.4\times10^{-20}$ and median $0.25$.
These failures cannot be used as rank lower bounds.

\par}

{\subsection{Dense random targets and mixtures of degrees}
\label{app:extra_random}

For scalar homogeneous targets with $d$ variables and $k\ge3$,
let $N=\binom{d+k-1}{k}$.
Outside the known defective cases, complex-generic ranks are
\[
 r_{\mathrm{tied}}^{\mathrm{gen}}=\lceil N/d\rceil,
 \qquad
 r_{\mathrm{untied}}^{\mathrm{gen}}
 =\lceil N/(2d-1)\rceil
\]
by the Alexander--Hirschowitz theorem and the tangential-Veronese
result~\citep{alexander1995polynomial,abo2018tangential}.
A defective case is one in which the dimension count
underestimates the required generic width.
For the tied family the count increases by one at
$(d,k)=(3,4),(4,4),(5,3),(5,4)$;
for the untied family it increases by one at
$(d,k)=(3,3),(4,3),(5,3)$.

We sample coefficients independently from a Gaussian in an
apolar-orthonormal basis and fix these predictions before fitting.
The observed recovery width is the smallest tested width reaching
relative loss below $10^{-10}$ in 32 restarts.
The complex lower threshold applies almost surely to these
continuously sampled real targets; it need not be an attainable
real width.
A real near-zero fit at that width supplies numerical recovery
evidence.
A fit only at the next width leaves higher real rank,
an ill-conditioned limit, and optimizer failure unresolved.

\begin{table}[t]
\centering\small
\caption{Dense random targets: observed recovery widths and
complex-generic predictions.
``Above'' marks recovery only above the prediction and does
not certify a larger real rank.}
\label{tab:random_dense}
\begin{tabular}{@{}cccccccl@{}}
\toprule
$d$ & $k$ & Target & $N$
& Tied obs./pred. & Untied obs./pred.
& Obs. ratio & Above prediction \\
\midrule
2 & 5 & 0 & 6  & 3/3   & 3/2 & 1.00 & Untied \\
2 & 5 & 1 & 6  & 4/3   & 2/2 & 2.00 & Tied   \\
2 & 6 & 0 & 7  & 4/4   & 3/3 & 1.33 & ---    \\
2 & 6 & 1 & 7  & 5/4   & 3/3 & 1.67 & Tied   \\
3 & 3 & 0 & 10 & 4/4   & 3/3 & 1.33 & ---    \\
3 & 3 & 1 & 10 & 4/4   & 3/3 & 1.33 & ---    \\
3 & 4 & 0 & 15 & 6/6   & 3/3 & 2.00 & ---    \\
3 & 4 & 1 & 15 & 7/6   & 4/3 & 1.75 & Both   \\
4 & 3 & 0 & 20 & 6/5   & 4/4 & 1.50 & Tied   \\
4 & 3 & 1 & 20 & 6/5   & 4/4 & 1.50 & Tied   \\
4 & 4 & 0 & 35 & 11/10 & 6/5 & 1.83 & Both   \\
4 & 4 & 1 & 35 & 10/10 & 6/5 & 1.67 & Untied \\
5 & 3 & 0 & 35 & 8/8   & 5/5 & 1.60 & ---    \\
5 & 3 & 1 & 35 & 8/8   & 5/5 & 1.60 & ---    \\
\bottomrule
\end{tabular}
\end{table}

Table~\ref{tab:random_dense} recovers several nontrivial
corrected counts.
The ternary cubic requires an untied prediction of three,
rather than the naive two.
At $(d,k)=(5,3)$, both predictions increase, to tied width
eight and untied width five; both are recovered on two targets,
with best losses at least $2.5\times10^{-4}$ one channel below
and below $10^{-15}$ at the predictions.
For tied $(3,4)$ target zero, losses below and at width six
are $1.47\times10^{-2}$ and $9.86\times10^{-19}$.
Observed tied/untied width ratios range from one to two;
the factor-$k$ separation of the monomial is a structured
example rather than a generic prediction.

\paragraph{Mixtures of degrees.}
We also fit independent random components at degrees $K$,
with shared directions but a free coefficient for each
channel and degree.
With $N_{\mathrm{tot}}=\sum_{k\in K}\binom{d+k-1}{k}$,
parameter counting gives the reference widths
\[
 m_t^{\mathrm{count}}
 =\left\lceil\frac{N_{\mathrm{tot}}}{d-1+|K|}\right\rceil,
 \qquad
 m_u^{\mathrm{count}}
 =\left\lceil\frac{N_{\mathrm{tot}}}{2d-2+|K|}\right\rceil.
\]
These are dimension-count benchmarks, not proved generic-rank
formulas for the shared-direction model.
The relaxation also differs from a fixed analytic gate
because its coefficients are free across degrees.

\begin{table}[t]
\centering\small
\caption{Exploratory degree-mixture fits.
Each entry is recovered width (dimension-count benchmark).
Paired entries correspond to two independently sampled targets.}
\label{tab:random_mixed}
\begin{tabular}{@{}ccccc@{}}
\toprule
$d$ & $K$ & $N_{\mathrm{tot}}$ & Tied & Untied \\
\midrule
2 & $\{2,3,4,5\}$ & 18
& $5\,(4),\ 4\,(4)$ & $4\,(3),\ 3\,(3)$ \\
3 & $\{2,3\}$ & 16
& $5\,(4),\ 4\,(4)$ & $4\,(3),\ 3\,(3)$ \\
3 & $\{2,3,4\}$ & 31
& $7\,(7)$ & $5\,(5)$ \\
4 & $\{2,3\}$ & 30
& $7\,(6),\ 7\,(6)$ & $4\,(4),\ 5\,(4)$ \\
\bottomrule
\end{tabular}
\end{table}

The untied recovered width is smaller in all seven comparisons
in Table~\ref{tab:random_mixed}.
The mixture with the largest coefficient dimension recovers
both counting benchmarks.
Recovery above a benchmark is unresolved and is not attributed
to a specific real rank or optimization mechanism.

\par}

{\subsection{Distillation of trained gated blocks}
\label{app:extra_distillation}

A frozen random tanh network $\mathbb R^8\to\mathbb R^8$
with hidden width 64 supplies a common target.
We train three teachers with 192 parameters each, excluding
biases: untied SwiGLU width eight, tied SwiGLU width twelve,
and a tanh MLP width twelve.
Their training relative MSEs on the common target are
$0.295$, $0.172$, and $0.112$, respectively.
Tied SwiGLU uses $\operatorname{SiLU}(z)z$, preserving the
untied model's gate while imposing $v=u$.

For each teacher and $m\in\{2,3,4,6,8,10,12\}$, we fit
untied width $m$, tied width $m$, and approximately
parameter-matched tied width $1.5m$ (rounded when necessary).
With input and output dimension eight, an untied channel has
24 parameters and a tied channel 16.
Counts therefore match exactly for the even base widths in
Table~\ref{tab:distillation}; odd-width rounding is approximate.
Inputs follow $\mathcal N(0,I_8)$, with 4,000 training and
4,000 held-out samples.
L-BFGS runs for up to 6,000 iterations.
The study contains $3\times7\times3\times3=189$ fits;
the table reports median relative held-out MSE over three
seeds at the retained representative widths.
Recovery uses a descriptive threshold of $10^{-3}$.

\begin{table}[t]
\centering\small
\caption{Trained-block distillation.
Ratios are tied/untied errors; values above one favor untied.
Ratios are retained at their reported precision and can differ
slightly from ratios of rounded losses.}
\label{tab:distillation}
\setlength{\tabcolsep}{4pt}
\begin{tabular}{@{}lcrrrrr@{}}
\toprule
Teacher & $m$ & Untied $m$ & Tied $m$ & Ratio
& Tied $1.5m$ & Ratio \\
\midrule
Untied & 4
& $4.02\!\times\!10^{-1}$ & $3.96\!\times\!10^{-1}$ & 0.99
& $2.29\!\times\!10^{-1}$ & 0.57 \\
Untied & 8
& $1.03\!\times\!10^{-1}$ & $1.16\!\times\!10^{-1}$ & 1.13
& $8.70\!\times\!10^{-4}$ & 0.01 \\
Untied & 10
& $1.37\!\times\!10^{-4}$ & $7.51\!\times\!10^{-2}$ & 549
& $1.09\!\times\!10^{-3}$ & 7.94 \\
Untied & 12
& $8.11\!\times\!10^{-6}$ & $8.70\!\times\!10^{-4}$ & 107
& $6.11\!\times\!10^{-4}$ & 75.3 \\
Tied & 8
& $2.82\!\times\!10^{-1}$ & $2.99\!\times\!10^{-1}$ & 1.06
& $1.64\!\times\!10^{-1}$ & 0.58 \\
Tied & 12
& $9.30\!\times\!10^{-2}$ & $1.64\!\times\!10^{-1}$ & 1.76
& $6.47\!\times\!10^{-2}$ & 0.70 \\
Tanh & 8
& $3.45\!\times\!10^{-1}$ & $3.44\!\times\!10^{-1}$ & 1.00
& $2.51\!\times\!10^{-1}$ & 0.73 \\
Tanh & 12
& $1.81\!\times\!10^{-1}$ & $2.51\!\times\!10^{-1}$ & 1.38
& $1.53\!\times\!10^{-1}$ & 0.84 \\
\bottomrule
\end{tabular}
\end{table}

On the untied teacher, the first equal-width median recovery
is at width ten for untied students and twelve for tied students.
Under parameter matching, tied width twelve already reaches
the criterion at base width eight.
Untied students achieve lower errors once they enter their
recovery regime, but the tied family also recovers the target.
Across all widths, geometric-mean tied/untied ratios are
$1.12$ at equal width and $0.67$ under matching for the tied
teacher, and $1.06$ and $0.78$ for the tanh teacher.
These controls do not show a consistent untied advantage.

At width ten on the untied teacher, one of three untied seeds
fails to recover, with losses $1.4\times10^{-4}$,
$1.2\times10^{-5}$, and $7.5\times10^{-2}$;
all three recover at width twelve.
A separate six-restart diagnostic at untied width eight
obtains best loss $2.75\times10^{-3}$, with other losses
$0.07$--$0.15$.
The teacher parameters themselves give an exact width-eight
representation, so these residuals are optimization failures.
This experiment tests recovery and precision for imperfect
trained teachers; it does not establish independent
architectural width lower bounds.

\paragraph{Norm continuation.}
For tied SwiGLU width twelve fitted to the untied width-eight
teacher, the best of three paths with a relaxing norm penalty
reaches held-out relative MSE $3.0\times10^{-4}$ at squared
parameter norm $\|\theta\|^2\approx180$, approximately five
times its initial value.
One path stalls near $7\times10^{-2}$.
No divergence is observed over the tested range.
The result concerns finite-error approximation under a fixed
parameterization; it proves neither closedness nor exact
representability and does not bound each Taylor coefficient
separately.
This raw squared parameter norm is distinct from the homogeneous
atom sizes used in the coefficient-space diagnostics.

\par}

{\subsection{Adam-trained gated scalar regression}
\label{app:extra_adam}

We additionally compare tied and untied SiLU-gated scalar blocks
at $d=8$ on two four-channel teachers.
The asymmetric teacher is
$\sum_j a_j\operatorname{SiLU}(p_j^\top x)(q_j^\top x)$
with independent standard Gaussian directions $p_j,q_j$ and
weights $a_j$; the tied teacher sets $q_j=p_j$.
Here the tied student is the matched-gate specialization
$\operatorname{SiLU}(u^\top x)(u^\top x)$.
Training uses 20,000 examples, 5,000 held-out examples,
Adam with learning rate $3\times10^{-3}$, batch size 256, and a
cosine schedule for 4,000 steps, and four seeds.
Table~\ref{tab:adam_budget} reports median relative test MSE.

\begin{table}[t]
\centering\small
\caption{Adam-trained scalar regression with approximately
matched total parameters. Counts are untied/tied;
the tied models have slightly larger budgets.
Bold marks the smaller error within each teacher and row.}
\label{tab:adam_budget}
\begin{tabular}{@{}c|rr|rr@{}}
\toprule
Parameters
& \multicolumn{2}{c|}{Asymmetric teacher}
& \multicolumn{2}{c}{Tied teacher} \\
Untied/tied & Untied & Tied & Untied & Tied \\
\midrule
17/18
& 0.3162 & \textbf{0.2580}
& 0.3808 & \textbf{0.1899} \\
34/36
& \textbf{0.0843} & 0.1049
& 0.0178 & \textbf{0.0014} \\
51/54
& \textbf{0.0322} & 0.0387
& 0.0045 & \textbf{0.0013} \\
68/72
& \textbf{0.0188} & 0.0370
& 0.0010 & \textbf{0.0006} \\
102/108
& 0.0088 & \textbf{0.0052}
& 0.0009 & \textbf{0.0006} \\
136/144
& 0.0062 & \textbf{0.0024}
& 0.0008 & \textbf{0.0004} \\
\bottomrule
\end{tabular}
\end{table}

The untied student has lower error on the asymmetric teacher
at the intermediate budgets of 34--68 parameters, whereas the tied student is more accurate at the
smallest budget and at the two largest budgets.
On the tied teacher, the tied student is more accurate at every budget.
Under approximately matched total parameters, the untied advantage is
therefore limited to a range of budgets on the asymmetric teacher.

%% file: figures/expressivity2/table_scope.tex
\begin{tabular}{@{}llcccc@{}}
\toprule
Target & Architecture & $m=1$ & $m=2$ & $m=3$ & $m=4$ \\
\midrule
$x_1^{k-1} x_2$ (role-asym.) & tied & $4.0\!\times\!10^{-1}$ & $2.0\!\times\!10^{-8}$ & $2.3\!\times\!10^{-10}$ & $1.2\!\times\!10^{-13}$ \\
 & untied (det.-val.) & $<\!10^{-15}$ & $6.3\!\times\!10^{-15}$ & $2.0\!\times\!10^{-15}$ & $<\!10^{-15}$ \\
 & multi-leg oracle & $1.2\!\times\!10^{-13}$ & $6.4\!\times\!10^{-14}$ & $<\!10^{-15}$ & $<\!10^{-15}$ \\
\midrule
$x_1 x_2 \cdots x_k$ (square-free) & tied & $6.1\!\times\!10^{-1}$ & $3.3\!\times\!10^{-1}$ & $1.4\!\times\!10^{-1}$ & $1.2\!\times\!10^{-15}$ \\
 & untied (det.-val.) & $3.3\!\times\!10^{-1}$ & $<\!10^{-15}$ & $<\!10^{-15}$ & $<\!10^{-15}$ \\
 & multi-leg oracle & $<\!10^{-15}$ & $<\!10^{-15}$ & $<\!10^{-15}$ & $<\!10^{-15}$ \\
\bottomrule
\end{tabular}